\documentclass[acmsmall, screen, nonacm]{acmart}

\usepackage{booktabs} %
\usepackage{algorithmic}
\usepackage{multirow}
\usepackage{xcolor}
\usepackage{pifont}
\usepackage{array}
\usepackage{amsmath}
\usepackage{subcaption}
\usepackage{xcolor,colortbl}
\usepackage{wasysym}
\usepackage{soul} %
\usepackage{graphicx}

\usepackage[ruled]{algorithm2e} %

\begin{document}

\setcopyright{none}

\title[InterAct: Dataset of Interactive Activities between Two People in Daily Scenarios]{InterAct: A Large-Scale Dataset of Dynamic, Expressive and Interactive Activities between Two People in Daily Scenarios}

\author{Leo Ho}
\authornote{Both authors contributed equally to this research.}
\email{leoh@connect.hku.hk}
\orcid{0009-0006-4373-2289}
\affiliation{%
    \institution{The University of Hong Kong}
    \city{Hong Kong}
    \country{Hong Kong}
}
\affiliation{%
    \institution{Centre for Transformative Garment Production}
    \city{Hong Kong}
    \country{Hong Kong}
}

\author{Yinghao Huang}
\authornotemark[1]
\email{yhuang2@gbu.edu.cn}
\orcid{0000-0002-8871-5128}
\affiliation{%
    \institution{Great Bay University}
    \city{Dongguan}
    \country{China}
}
\affiliation{%
    \institution{Dongguan Key Laboratory for Intelligence and Information Technology}
    \city{Dongguan}
    \country{China}
}

\author{Dafei Qin}
\affiliation{%
  \institution{The University of Hong Kong}
  \city{Hong Kong}
  \country{Hong Kong}
}
\email{qindafei@connect.hku.hk}
\orcid{0009-0001-4992-4760}

\author{Mingyi Shi}
\affiliation{%
  \institution{The University of Hong Kong}
  \city{Hong Kong}
  \country{Hong Kong}
}
\email{mingyis@connect.hku.hk}
\orcid{0000-0002-5180-600X}

\author{Wangpok Tse}
\affiliation{%
    \institution{The University of Hong Kong}
    \city{Hong Kong}
    \country{Hong Kong}
}
\email{crazytse@connect.hku.hk}
\orcid{0009-0005-3836-4875}

\author{Wei Liu}
\affiliation{%
    \institution{Shandong University}
    \city{Jinan}
    \state{Shandong}
    \country{China}
}
\email{202100130071@mail.sdu.edu.cn}
\orcid{0009-0004-7554-1730}

\author{Junichi Yamagishi}
\affiliation{%
    \institution{National Institute of Informatics}
    \city{Tokyo}
    \country{Japan}
}
\email{jyamagis@nii.ac.jp}
\orcid{0000-0003-2752-3955}

\author{Taku Komura}
\affiliation{%
    \institution{The University of Hong Kong}
    \city{Hong Kong}
    \country{Hong Kong}
}
\affiliation{%
    \institution{Centre for Transformative Garment Production}
    \city{Hong Kong}
    \country{Hong Kong}
}
\email{taku@cs.hku.hk}
\orcid{0000-0002-2729-5860}

\begin{abstract}
We address the problem of accurate capture of interactive behaviors between two people in daily scenarios. Most previous works either only consider one person or solely focus on conversational gestures of two people, assuming the body orientation and/or position of each actor are constant or barely change over each interaction. In contrast, we propose to simultaneously model two people's activities, and target objective-driven, dynamic, and semantically consistent interactions which often span longer duration and cover bigger space. To this end, we capture a new multi-modal dataset dubbed InterAct, which is composed of 241 motion sequences where two people perform a realistic and coherent scenario for one minute or longer over a complete interaction. For each sequence, two actors are assigned different roles and emotion labels, and collaborate to finish one task or conduct a common interaction activity. The audios, body motions, and facial expressions of both persons are captured. InterAct contains diverse and complex motions of individuals and interesting and relatively long-term interaction patterns barely seen before. We also demonstrate a simple yet effective diffusion-based method that estimates interactive face expressions and body motions of two people from speech inputs. Our method regresses the body motions in a hierarchical manner, and we also propose a novel fine-tuning mechanism to improve the lip accuracy of facial expressions. To facilitate further research, the data and code is made available at \url{https://hku-cg.github.io/interact/}.
\end{abstract}



\maketitle

{
\renewcommand\thefootnote{}
\footnotetext{\copyright~Owner/Author 2025. This is the author's version of the work. It is posted here for your personal use. Not for redistribution. The definitive Version of Record was published in \textit{Proc. ACM Comput. Graph. Interact. Tech., Vol. 8, No. 4, Article 53 (August 2025)}, \url{https://doi.org/10.1145/3747871}.}
}

\section{Introduction}
The interaction between two people during conversation is multi-modal: humans communicate not only through speech but also through facial expressions, body gestures, and the space between the two. 
The interaction behaviors can also be of various temporal and spatial scales. For example, two people can stand still while talking about many topics casually, and they can also walk over a large area to collaborate on a long-duration task. Different kinds of interaction between two people are ubiquitous in daily life.  

Most previous works that produce facial motion and body gestures from speech do not fully consider such multi-modal nature of conversation. They typically assume that each person moves in a small range, ignoring possible large-scale and long-term interactive activities between two people. Existing work mainly regress the speech data to the face \cite{fan2022faceformer, VOCA2019, karras2017audio, stan2023facediffuser} and body motion \cite{pang2023bodyformer, alexanderson2023listen} or condition the generative model to the speech contents and a prompt that describes the motion \cite{tevet2022human, xu2023inter, liu2022beat}. 
The works are mostly for motion synthesis of a single agent. 
Although some recent works estimate full body and facial motion of two people from speech \cite{ng2024audio2photoreal}, the setup is limited to subjects standing upright with little dynamic interactions between the two.

\begin{figure}
    \centering
    \includegraphics[width=0.99\linewidth]{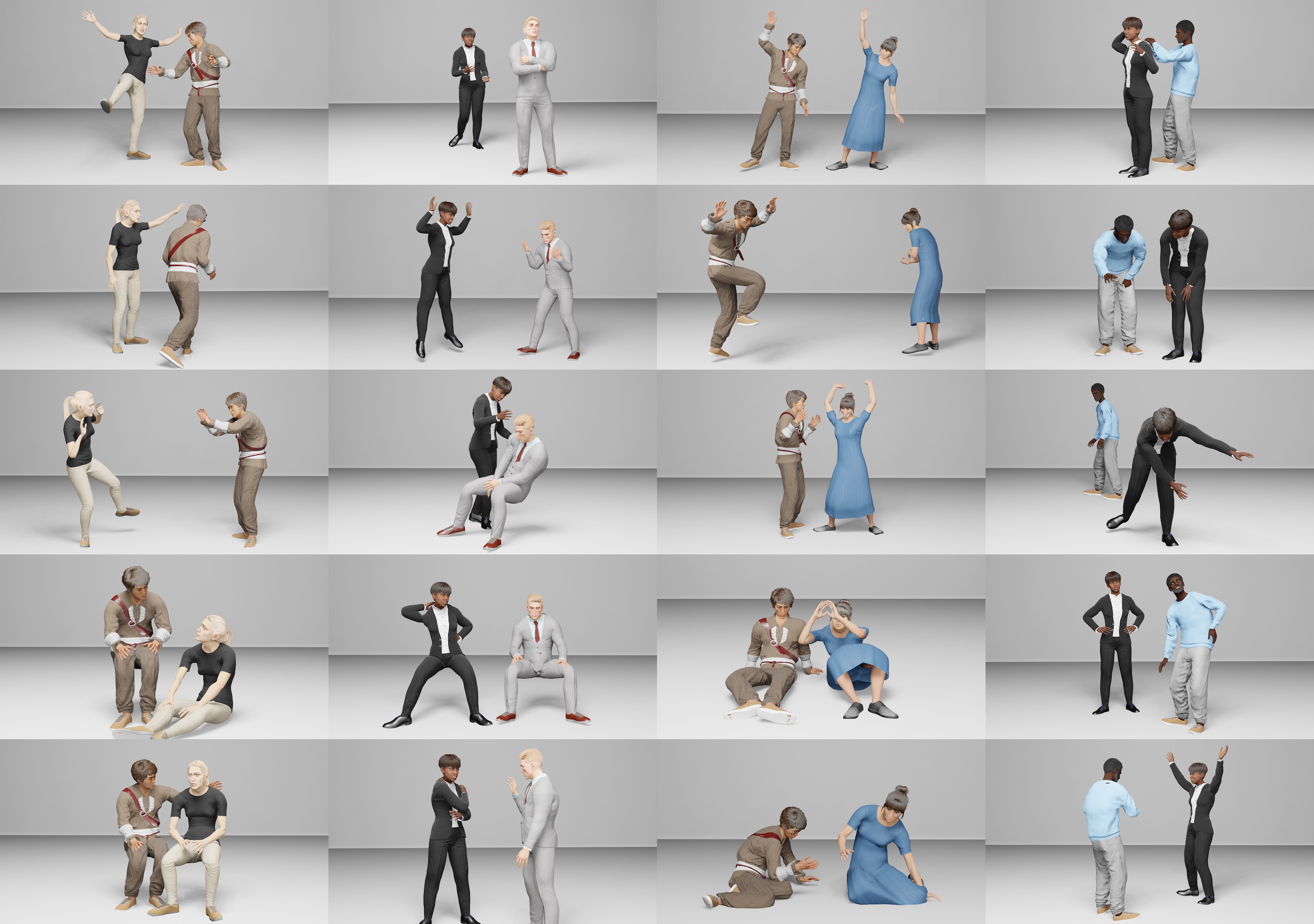}
    \caption{We capture and model realistic interactions between two people in daily life. Our dataset includes a diverse range of novel and dynamic interactions as shown above, showing humans in various poses and engaged in different activities.
    }    
    \label{fig:teaser}
\end{figure}

In this paper, we present a novel multi-modal motion capture dataset titled InterAct that simultaneously captures body marker motion captures, facial mesh animations, and speech audio of two interacting actors. To prepare the dataset, scenarios that vary in emotions and inter-personal relationships are systematically categorized and acted out by the actors, allowing actors to explore a wide variety of topics, ranging from daily occurrences to supernatural phenomena. For every scenario, the actors perform a one-minute act with emotional speech, vivid expressions, body gestures, and possibly dynamic movements. See Fig. \ref{fig:teaser} for examples. Unlike previous two-person datasets which focuses on short clips of body interactions, or gesture variations during a long conversation, our dataset captures the full behaviors of two people under different scenarios, collaboration tasks, emotional contexts, and relationships. This results in a much more comprehensive look into human behavior that benefits downstream tasks such as two-person motion generation, motion estimation, as well as improving existing models with human behavior grounding and artificial empathy.

We include a baseline method trained on our newly-captured dataset that jointly regresses realistic interactive body motions and facial blendshape parameters for two people from their speech and action labels. We show that our system trained on InterAct produces interesting interactions of two agents from speech, including vivid facial expressions and dynamic body motions. Finally, we provide rendered body and facial visualizations with audio on fully rigged characters on all captures of our dataset. All data, code, and trained models will be made public upon paper acceptance.

\section{Related Work}

\subsection{Human Motion Capture and Synthesis}

The domain of curating single-person human motion has been widely explored. Human3.6M \cite{ionescu2013human3} is one of the first large-scale motion datasets that capture accurate joint positions and angles for various daily life scenarios. Trinity Speech-Gesture \cite{ferstl2018investigating} captures a dataset of spontaneous speech and body markers from a single actor. AMASS \cite{mahmood2019amass} integrates prior marker-based motion datasets into a SMPL-based unified parameterization. TalkSHOW \cite{yi2023generating} constructs a multi-modal dataset consisting of upper body, face, hand, and speech fitted from Internet videos. ZeroEGGS \cite{ghorbani2023zeroeggs} records the monologue of an English-speaking female actor performing in various styles. Motorica \cite{alexanderson2023listen} captures performers dancing to different genres of music. BEAT \cite{liu2022beat} utilizes MoCap to capture full-body monologue data, which is enriched with emotional and semantic labels. HumanML3D \cite{guo2022generating} and Motion-X \cite{lin2023motionx} annotate textual descriptions to motion sequences to enable the learning of joint motion-language embeddings.

For simultaneous motions of two or more people, CMU Panoptic Dataset~\cite{Joo_2017_TPAMI} captures RGB-D multiview data from multiple people interacting and reconstructs body poses and facial landmarks. TalkHands~\cite{lee2019talking} and~\cite{ng2024audio2photoreal} document two actors engaged in dyadic conversations using motion capture. ConvoFusion~\cite{mughal2024convofusion} records a group of 5 people chatting together. On the other hand, Hi4D~\cite{yin2023hi4d} and Inter-X~\cite{xu2023inter} capture various physical interactions of two actors while adding detailed textual descriptions. However, none of the previous methods capture both conversations and interactions of two people. The dynamic interactions of two actors over a large capture space remain relatively unexplored. See Tab. \ref{tab:dataset_comp} for a detailed numerical comparison between our dataset and existing ones. 

As for human motion synthesis, PCA has been a foundational tool for reducing the dimensionality of motion data~\cite{howe1999bayesian, safonova2004synthesizing, chai2005performance}. To address the limitations of linear methods like PCA, kernel-based approaches have been introduced to model the nonlinear characteristics inherent in human motion.~\cite{mukai2005geostatistical, grochow2004style, levine2012continuous}. Classical methods developed in graphics are mostly rule-based \cite{levine2009real, neff2008gesture}; a set of template motions are prepared and launched according to speech or contextual features, which result in repeating similar movements despite the wide range of topics spoken by the character. To produce natural and responsive human motion, motion matching techniques have been developed to synthesize movements from reference databases in real-time~\cite{Habibie_gesturematching22, van2014motion, clavet2016motion}.  Recent advancements have integrated learnable functions and categorical codebooks to enhance the flexibility and accuracy of motion matching~\cite{holden2020learned, starke2024categorical}. The GENEA Challenge \cite{kucherenko2023genea} maintains a yearly benchmark for the speech-driven gesture generation task, with many methods being network-based.

Neural networks, recognized for their capacity as general function approximators, have been extensively applied in motion synthesis. Early works by~\cite{taylor2009factored} utilized conditional Restricted Boltzmann Machines (cRBM) for predicting subsequent poses in locomotion sequences. Building upon this,~\cite{fragkiadaki2015recurrent} proposed an Encoder-Recurrent-Decoder (ERD) network that employs Long Short-Term Memory (LSTM) units to model human dynamics, capturing temporal dependencies in motion data. More recent approaches, such as the Phase-Functioned Neural Network (PFNN)~\cite{holden2017phase, starke2022deepphase}, have demonstrated the ability to generate high-quality, real-time character animations by explicitly modeling the phase information of cyclic motions. Generative models such as normalizing flows~\cite{DBLP:journals/cgf/AlexandersonHKB20,valle2021transflower}, diffusion models~\cite{tevet2022human, chen2024taming} are leveraged for better diversity and generalization. 

In the domain of audio-driven gesture synthesis, early attempts directly map the speech features to body motion, which suffer from smoothing artifacts and lack of contextual motion~\cite{ginosar2019learning}. To mitigate these issues, methods based on motion matching~\cite{Habibie_gesturematching22}, generative models such as normalizing flows~\cite{DBLP:journals/cgf/AlexandersonHKB20,valle2021transflower}, variational Transformers~\cite{pang2023bodyformer} and diffusion models~\cite{alexanderson2023listen, ao2023gesturediffuclip} are developed. These methods better capture the weak correlations between speech and motion, leading to more natural and expressive animations. For motion stylization and control,~\citet{ghorbani2023zeroeggs} provide template motion to control the style,~\citet{ao2023gesturediffuclip} provide prompt as an additional input, and large language models are also used in recent literature~\cite{zhang2024semantic} to support semantic conditioned synthesis.

\subsection{Facial Expression Capture and Synthesis}

Many 2D and 3D datasets solely focus on the capture of speech and face. VOCASET \cite{VOCA2019} introduces a 12-person dataset totaling 29 minutes of scanned 3D oration performances. MEAD \cite{wang2020mead} builds a 2D talking-face dataset comprising 60 actors talking with varying levels of intensity and emotions. Many current audio-to-face models are trained on datasets comprising multiple data sources. EmoTalk \cite{Peng_2023_ICCV} integrates two 2D emotional audio-face datasets, RAVDESS \cite{livingstone2018ryerson} and HDTF \cite{zhang2021flow}, and reconstructs 3D meshes and blendshapes for talking face generation. Most recently, UniTalker \cite{fan2024unitalker} assembles existing and newly collected datasets to create a large-scale talking-face collection with 18.5 hours of data.

As for audio-driven facial animation systems,  Early works build mappings between visemes and talking shapes manually~\cite{ezzat1998miketalk, de2006facial}. Dimension reduction tools like PCA are deployed to reduce the complexity of such a mapping~\cite{kalberer2001face}. Hybrid methods bridge data-driven and artistic control, enabling expressive animation while maintaining accuracy~\cite{edwards2016jali}.

Recently, \citet{taylor2017deep} regresses the MFCC features to face rig parameters by temporal convolution (TCN).~\citet{fan2022faceformer} makes use of the Transformer structure to learn the subtle correlation between speech and lips. VAE has been applied to the facial animation sequence for dimension reduction~\cite{richard2021meshtalk, villanueva2022voice2face}. Learning on this latent manifold mitigates the effect of mode averaging and collapse, leading to more natural expressions like eye-blinking and subtle upper-face emotions~\cite{richard2021meshtalk,Xing_2023_CVPR}. 

Apart from learning accurate speech-to-mouth mapping, the subtle emotions humans can express and perceive is another challenge.~\citet{karras2017audio} introduces the emotion vector to which the emotion parameters are encoded during training.~\citet{Peng_2023_ICCV}  disentangle the emotion from the data using an emotion encoder.~\citet{EMOTE} learn emotional facial expressions using a temporal VAE structure; in addition to the speech, the emotion label is given as an input to control to produce rich expressions of the characters.
Further, they improve the quality of lip synchronization by using an image projection loss when learning the motion from videos.
Several methods make use of diffusion models to improve the variation of the expressions during speech~\cite{stan2023facediffuser}, with multi-modal conditions and classifier-free guidance to help provide more diverse and controllable expression details~\cite{zhao2024media2face}.
Most recently,.~\citet{xu2024vasa} employs a novel latent disentanglement scheme to develop an expressive latent space, enabling the generation of realistic 2D facial animations.

\subsection{Joint Synthesis of Body and Face from Speech}
Very recently, efforts have been made to synthesize both the facial expressions and body motions from the speech~\cite{yi2023generating,liu2023emage,ng2024audio2photoreal}.
SHOW~\cite{yi2023generating} trains separate VQ models for the face, body, and fingers to simultaneously control them through speech. 
EMAGE~\cite{liu2023emage} prepares Transformer-based controllers for different parts of the body to produce motion that includes global translation.~\citet{ng2024audio2photoreal} proposes a VQ and diffusion-based framework to individually train face and body motions which are unified to get photo-realistic whole-body rendering.

There has also been an increasing interest in producing motions based on two-person conversational scenarios
~\cite{ng2021body2hands,ng2022learning,Ng_2023_ICCV, pang2023bodyformer,ng2024audio2photoreal}. However, these methods only model a single person, omitting the interactions between the two people. Possible modalities such as the distance between characters, facial expression, and body pose correlations, are not considered as well.~\cite{liang2024intergen} and ~\cite{ghosh2024remos} model the two-person interaction by synthesizing one person's reaction given the another person's motion clip. However, these methods neglect the verbal information rapidly transferred between daily two-person scenarios.

\section{InterAct Dataset}

We first describe how we build our new two-person interaction dataset, which is our main contribution. Then we conduct statistical analysis on the newly captured dataset, and compare it with two closely related ones to demonstrate the diversity and quality of our dataset.

\subsection{Dataset Construction}

\begin{figure}[h]
    \centering
    \begin{subfigure}[h]{0.32\linewidth}
        \includegraphics[width=1\linewidth]{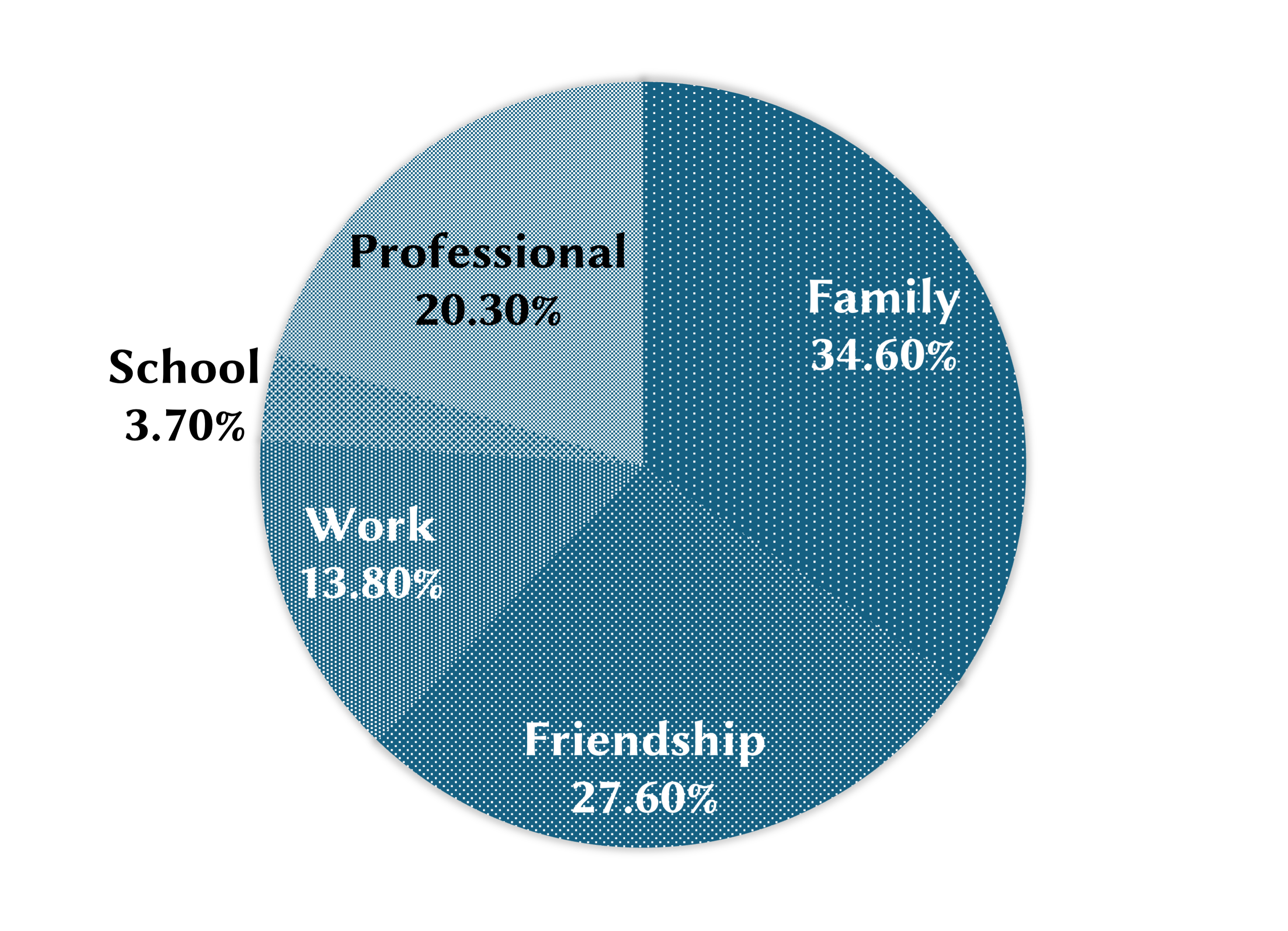}
        \caption{By meta-relationship}
    \end{subfigure}
    \begin{subfigure}[h]{0.32\linewidth}
        \includegraphics[width=1\linewidth]{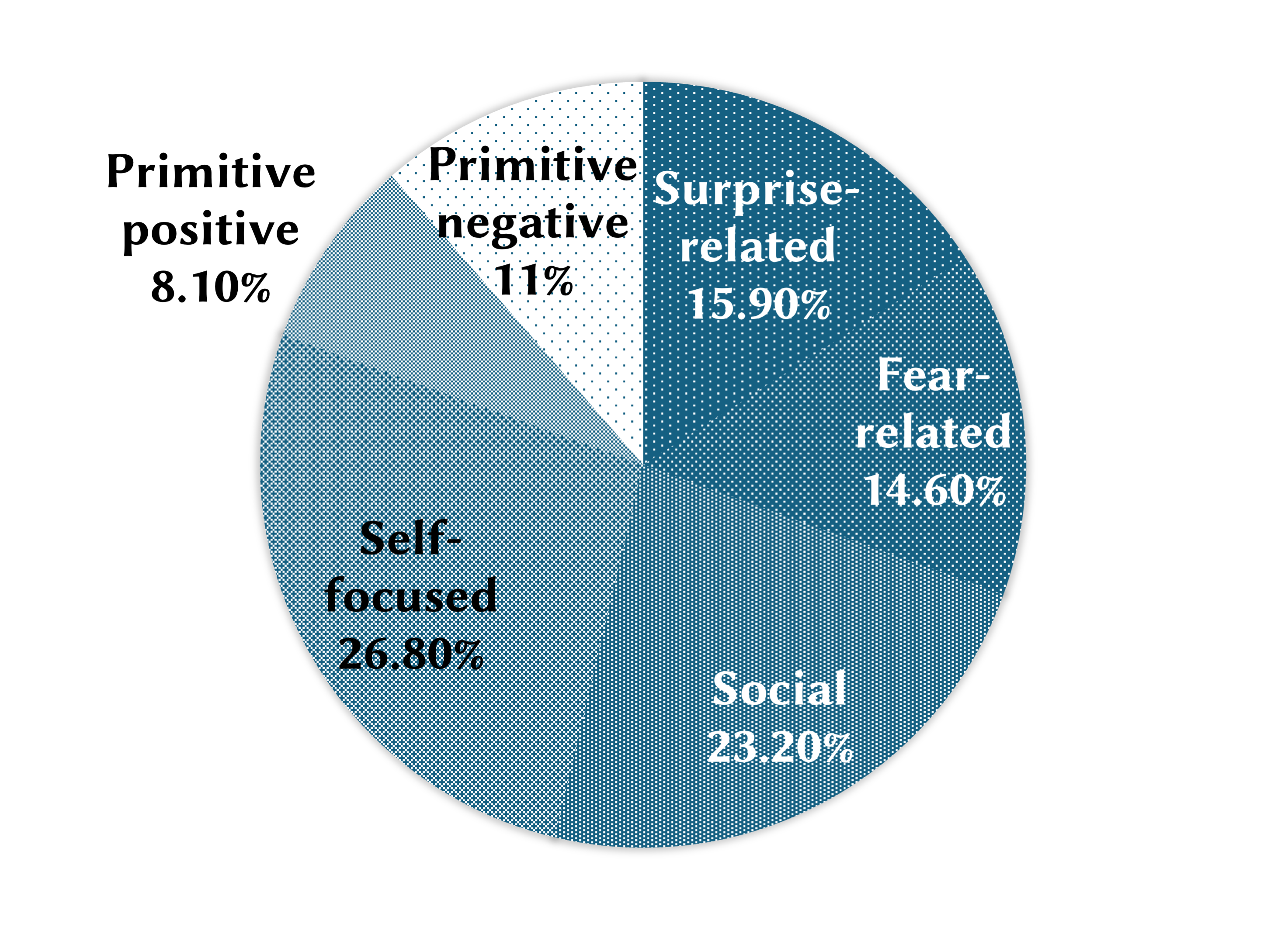}
        \caption{By meta-emotion}
    \end{subfigure}
    \begin{subfigure}[h]{0.32\linewidth}
        \includegraphics[width=1\linewidth]{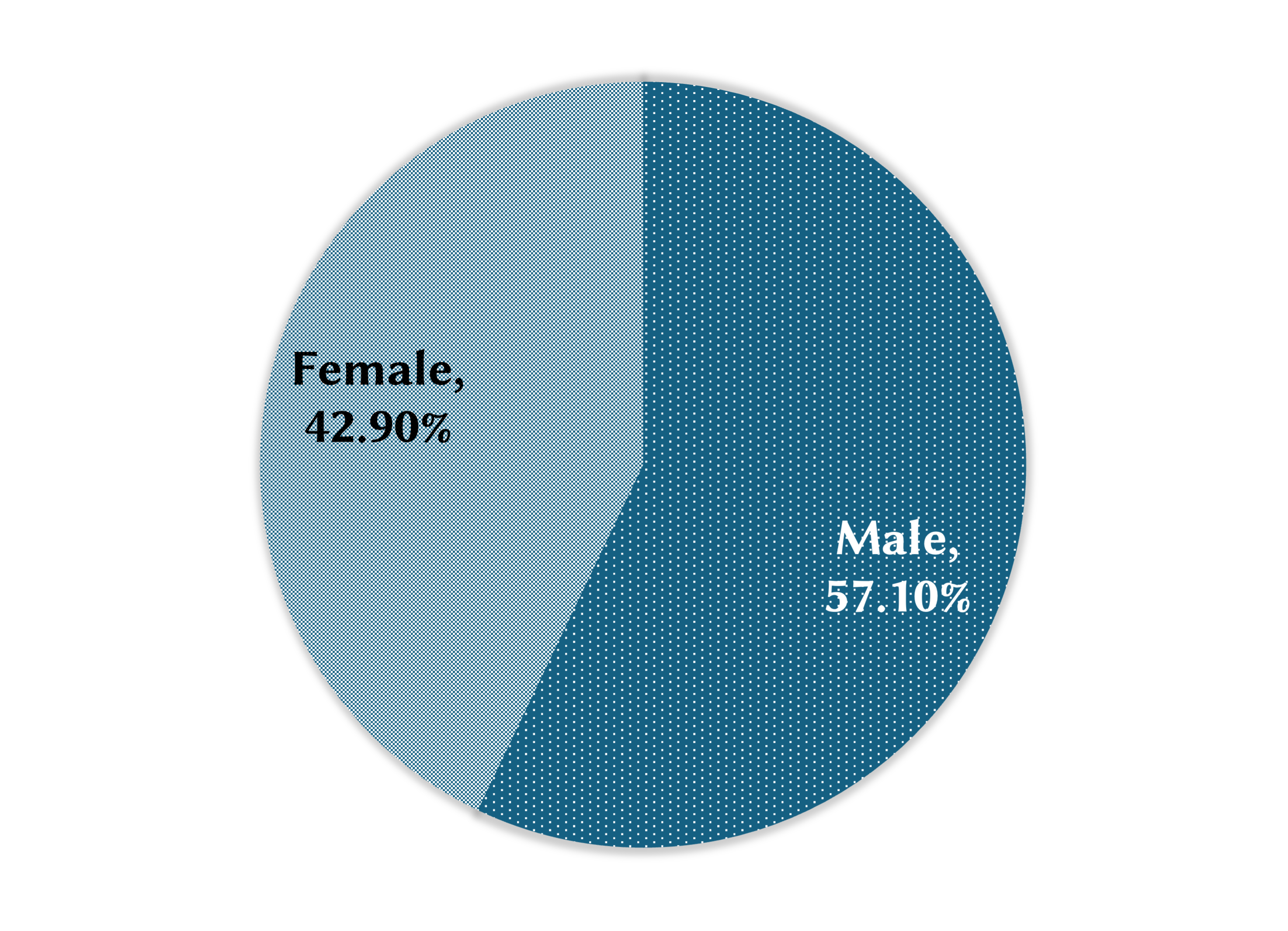}
        \caption{By gender}
    \end{subfigure}
    \caption{Breakdown of InterAct dataset by meta-relationships, meta-emotions, and gender. Fine-grained relationships and emotions given to actors as prompts are grouped into meta-relationships and meta-emotions for better visualization. For example, family includes sibling, cousin, in-law, parent, grandparent, and husband / wife.}
    \label{fig:pie_chart}
\end{figure}

We want the new dataset to be rich in relationships, emotions, and scenarios.  Settings such as family, work, friends, and school are captured in our dataset. Our selection covers distinct relationships, including but not limited to siblings, parent-child, doctor-patient, and teacher-student. The emotional spectrum of our dataset references the work in \cite{doi:10.1073/pnas.1702247114} and incorporates a modified subset of emotions that exist in distinct categories. %
These emotions include joy, adoration, interest, boredom, anxiety, among other, totaling 26 in all. 
The full list of emotions and relationships can be found in the supplementary.
By design each sequence in InterAct is labelled with emotion state, relationship and the genders of both actors. Here we show the distributions of all the motion sequences w.r.t. these factors in Fig. \ref{fig:pie_chart}. Overall, the distribution is quite even with the exception that the \textit{School} class in the \textit{Relationship} category is underrepresented.

\subsubsection{Data Capture}
\begin{figure}[t]
    \centering
    \includegraphics[width=1.0\linewidth]{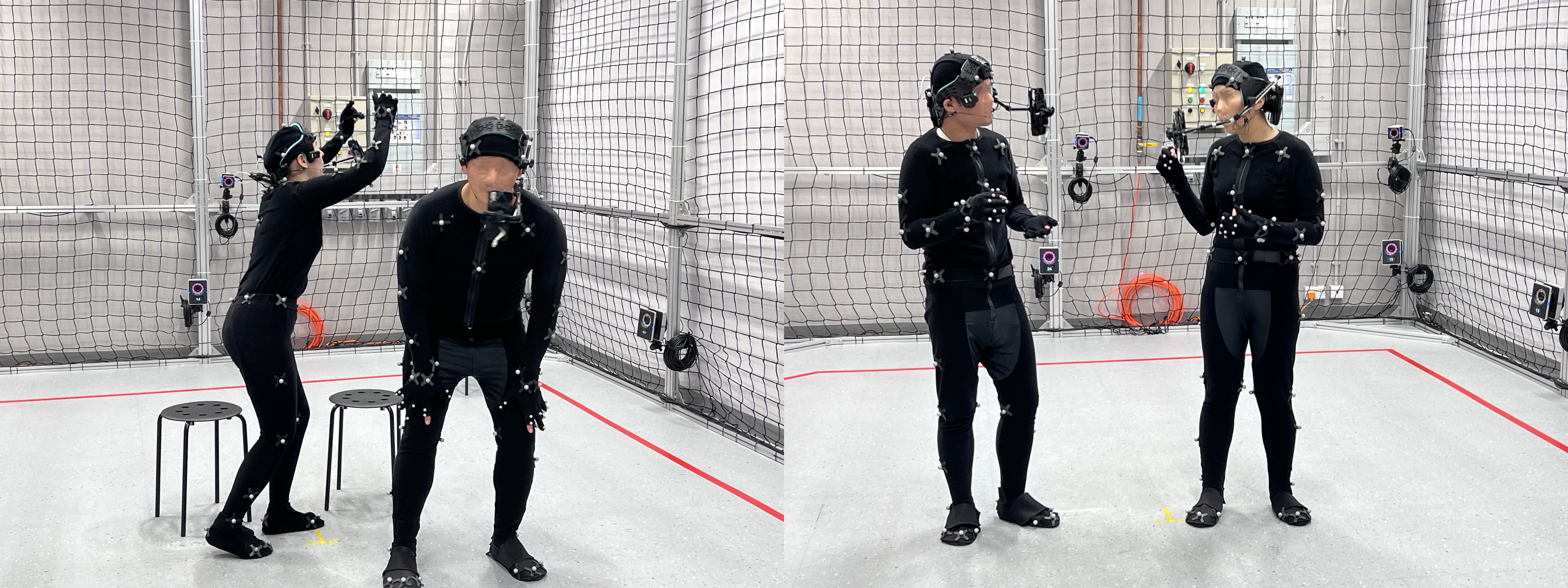}
    \caption{Two pairs of actors during capture sessions. Their audios, facial expressions and body motions are all recorded.
    }
    \label{fig:capture_actors}
\end{figure}

The capture system consists of several components, either worn by the actors or stationed around the capture space. To capture both actors' body motions, a 28-camera VICON optical MoCap system is positioned around the 5m x 5m space, with each camera located in one of three elevations. We place 53 body markers and 20 finger markers on each actor, and ask them to perform a Range of Motion exercise for actor calibration before capture starts. The front depth sensor and camera of an iPhone are used to capture the facial expressions of each actor. Prior to the capture, a mesh template of each actor's front face is registered by asking the actor to rotate her/his head while facing the camera. During the capture, an actor wears a head rig that holds one iPhone, two microphones (one active and one backup), and a power bank that also acts as a counterweight. A snapshot of two pairs of actors during data capture is shown in Fig. \ref{fig:capture_actors}.

As the motion and face capture systems are separate, they need to be temporally synced for accuracy. To this end, a wireless timecode generator is used to broadcast the clock signal of the VICON system to individual iPhones, syncing the timecode at the beginning of a shoot. Before and after each capture session, a script is triggered to synchronously start and stop the recording of both systems at a given timecode. This automatically ensures frame-level accuracy while requiring less post-processing work to sync the two sources together. During the capture, the live audio and face video of both actors are streamed to the control station for real-time monitoring, which ensures data quality. The capture is interrupted and restarted once any error is noticed.

\paragraph{Data Format}
Altogether we gather 241 sequences acted by 7 participants. For every sequence, the speeches of both actors are recorded in separated WAV files. Then VisualVoice \cite{gao2021visualvoice}, a face-conditioned speech separation method, is employed to extract the clean speech of each actor. The body motions are exported from VICON in BVH format, one file for each actor-sequence. Note that intricate hand movements are also captured. For the face, we first obtain a personalized facial mesh template alongside detailed vertices flows, then convert them into frame-by-frame facial meshes, and finally to the widely used ARKit format using the least-squares algorithm. We detail the facial conversion process in our supplementary. For better modeling and generating meaningful human motions, we compute the action mode in a heuristic way and assign a label $C$ among \{sit, walk, stand\} to each actor of every frame. These annotations will be released alongside the data.

\paragraph{Diversity} 
Best efforts have been undertaken to achieve the diversity of the dataset. We recruit 7 actors (4 males and 3 females) from varying age groups and nationalities. For every scenario, a pair of male and female actors perform impromptu for one minute or longer, without any interference or extra guidance apart from the given character setup and scenario description prompts. Two chairs are available to them as a prop, which they can freely arrange or use to simulate various real-life situations, such as the front seats of a vehicle, a classroom, or a doctor's office.

\paragraph{Additional Face Dataset}
The use of a close-proximity front-facing iPhone camera poses a minor challenge to face reconstruction due to lens distortion. To improve lip accuracy, an additional 1 hour of facial recording is captured where one actor reads sentences from the TIMIT database \cite{garofolo1993darpa} while directly facing a camera 0.6m away. Special care is taken to ensure the accuracy of the pronunciation and the lip shapes while recording. We convert the data into ARKit format for model fine-tuning, which will be discussed later. This additional dataset will be released alongside the main dataset to facilitate reproducibility.

\begin{table}[t]
    \caption{Comparison between our new dataset and existing ones. InterAct is the first to simultaneously capture audios, dynamic body motions, and facial meshes of two actors. It features relatively long-term and coherent interactive activities. The works listed here are: TalkSHOW\cite{yi2023generating}, ZeroEGGS\cite{ghorbani2023zeroeggs}, Motorica\cite{alexanderson2023listen}, BEAT\cite{liu2022beat}, Ng et al.\cite{ng2024audio2photoreal}, Inter-X \cite{xu2023inter}, Hi4D\cite{yin2023hi4d} and TalkHands \cite{lee2019talking}.}
    \centering
    \resizebox{\textwidth}{!}{
    \begin{tabular}{cc|cccccc|ccccccc}
        \toprule
        \#Actor & Name & Type  & \#Ppl. & \#Seq. & \#Joints  & Dura. & Avg.Len. & Audio & Hands & Face & \#Role & \#Emo. & Coher. & Dyna.\\        
        \midrule
       \multirow{4}{*}{1} & TalkSHOW & Mono  & 4 & 9,720  & 46  & 27.0h & 10.0s  & \checkmark & Coarse  & \checkmark & \ding{55} & \ding{55} & \ding{55} & \ding{55}\\   
       
       & ZeroEGGS & Mono & 1  & 67 & 75 & 2.3h  & 2.0m & \checkmark & Detailed  & \ding{55} & \ding{55} & \ding{55} & \ding{55} & \ding{55}\\

       & Motorica & Dance & 5  & 244 & 50  & 7.0h & 1.7m & \checkmark &  Detailed   & \ding{55} & \ding{55} & \ding{55} & \ding{55} & \checkmark\\        

       & BEAT & Mono & 30  & 2,508 & 75  & 76.0h & 1.8m & \checkmark & Detailed  & \checkmark & \ding{55} & 8 & \checkmark & \checkmark\\
       \hline 
       \multirow{5}{*}{2}& Ng et al. & Conver. & 4  & 48 & 56 
 & 8.0h & 4.9m & \checkmark & Detailed & \checkmark & \ding{55} & \ding{55} & \checkmark & \ding{55}\\        
        & Inter-X & Inter. & 89  & 11,388 & 54 & 18.8h  & 6.0s & \ding{55}  & \ding{55} & \ding{55} & \ding{55} & \ding{55} & \checkmark & \checkmark\\
        
       & Hi4D & Inter. & 20  & 100 & 24 & 6.1m & 3.6s &\ding{55} & \ding{55}  & \checkmark & \ding{55} & \ding{55} & \checkmark & \checkmark\\    
       
        & TalkHands & Conver. & 50  & 200 & 82 & 50.0h  & 10.0m &\checkmark & Detailed  & \ding{55} & \ding{55} & \ding{55} & \ding{55} & \ding{55}\\    

       \rowcolor[gray]{0.80}
       & InterAct & Conver. & 7  & 241 & 61  & 10.0h & 1.2m  & \checkmark  & Detailed & \checkmark & 25 & 26 & \checkmark & \checkmark\\
        \bottomrule
    \end{tabular}
    }
    \label{tab:dataset_comp}
\end{table}

\subsection{Dataset Analysis}
\label{sec:analysis}
In this section we first conduct statistical analysis on the newly captured dataset, and compare it with other speech-to-motion datasets. The comparison is shown in Table \ref{tab:dataset_comp}. Our dataset is the only one that captures both actors' body motions and facial expressions at the same time. In addition, our dataset contains more diverse and long-term interactions than other datasets.
To investigate the superiority of our dataset in dynamic and long-term interactions, we also perform a qualitative analysis on the newly captured motion sequences in our dataset, and compare it with two closely related ones: BEAT \cite{liu2022beat} and the one introduced in \cite{ng2024audio2photoreal}. Note that BEAT records one single actor's performance in each session, while \cite{ng2024audio2photoreal} records both actors primarily during conversation standing still.

\subsubsection{Dynamic and Long-term Interactions}

\begin{figure}[h]
    \centering
    \begin{subfigure}[b]{0.49\linewidth}
        \includegraphics[width=0.96\linewidth]{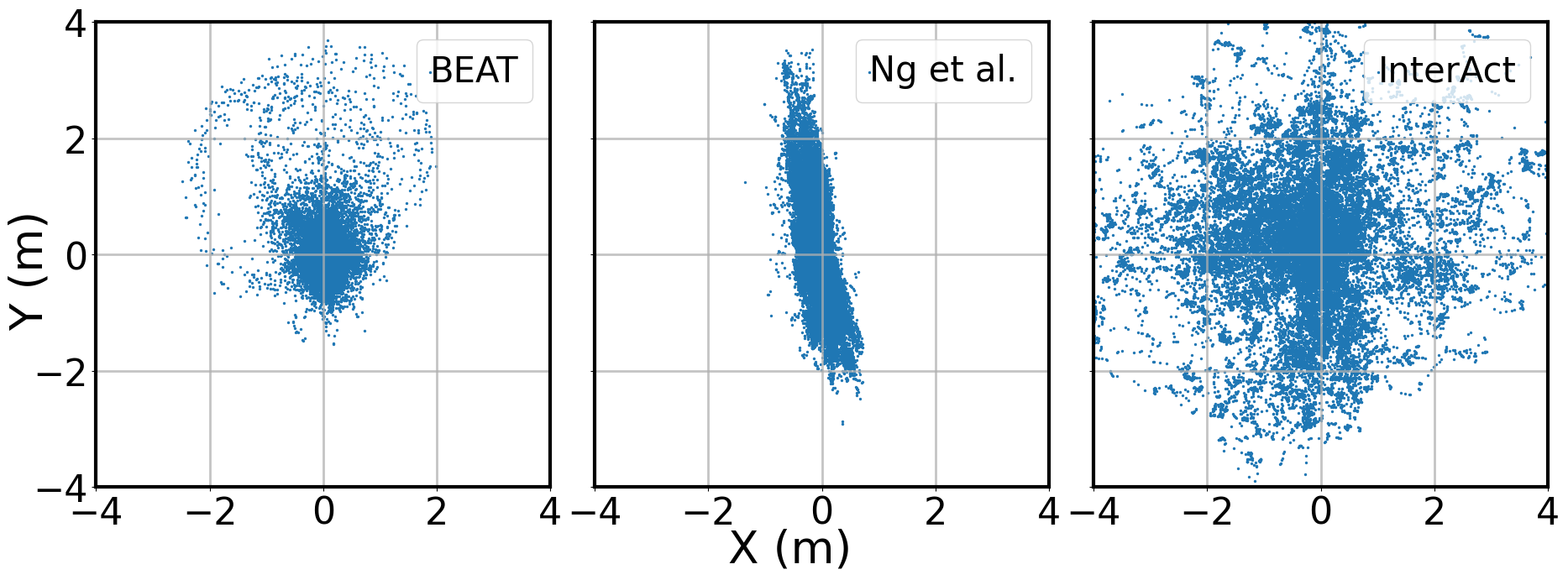}
        \caption{Scatter graphs of body positions projected on the ground plane with respect to the first frame.}
        \label{fig:subfig_a}
    \end{subfigure}\hfill
    \begin{subfigure}[b]{0.49\linewidth}
        \includegraphics[width=0.338\linewidth]{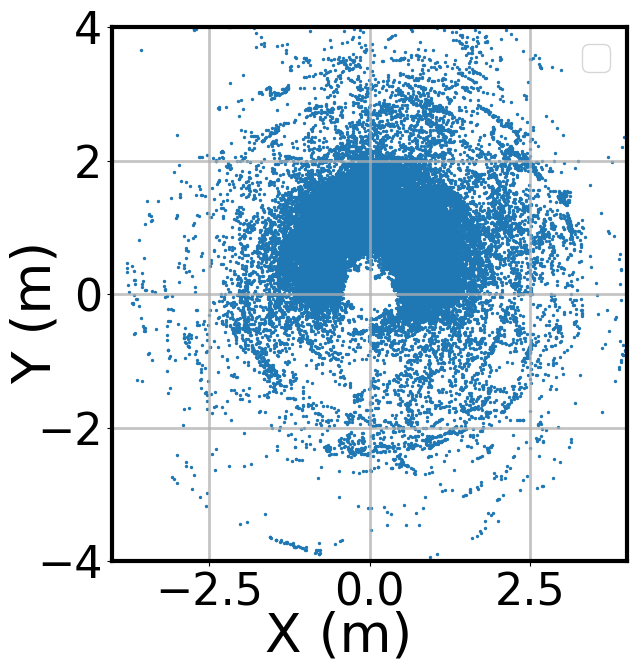}%
        \includegraphics[width=0.355\linewidth]{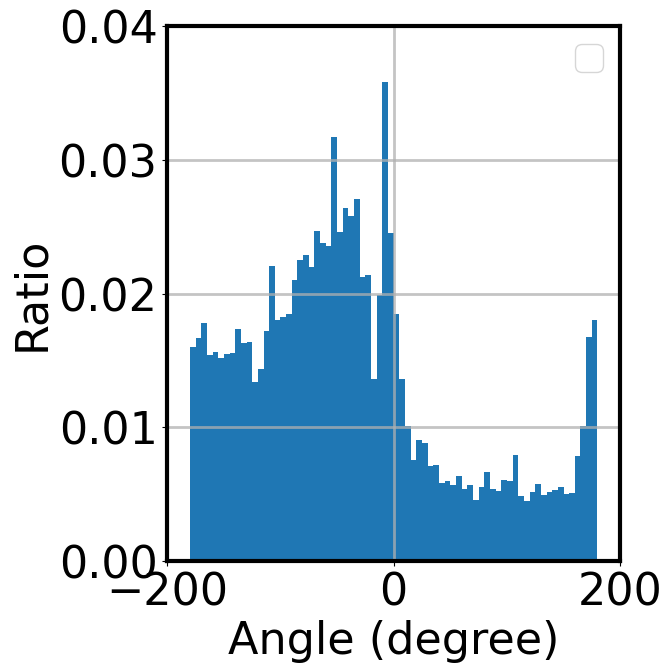}%
        \includegraphics[width=0.355\linewidth]{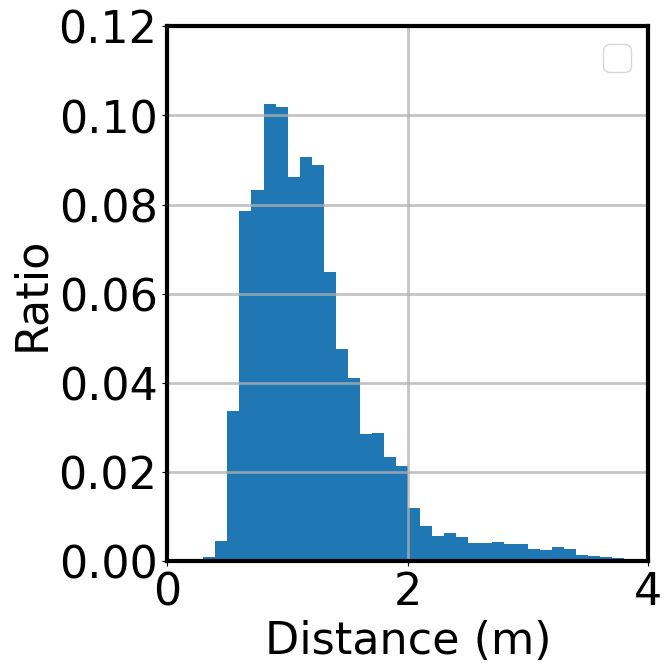}
        \caption{Spatial relations of the two subjects; scatter graph of relative positions, distribution of angle/distance.}
        \label{fig:subfig_b}
    \end{subfigure}

    \begin{subfigure}[b]{0.49\linewidth}
        \includegraphics[height=0.36\linewidth]{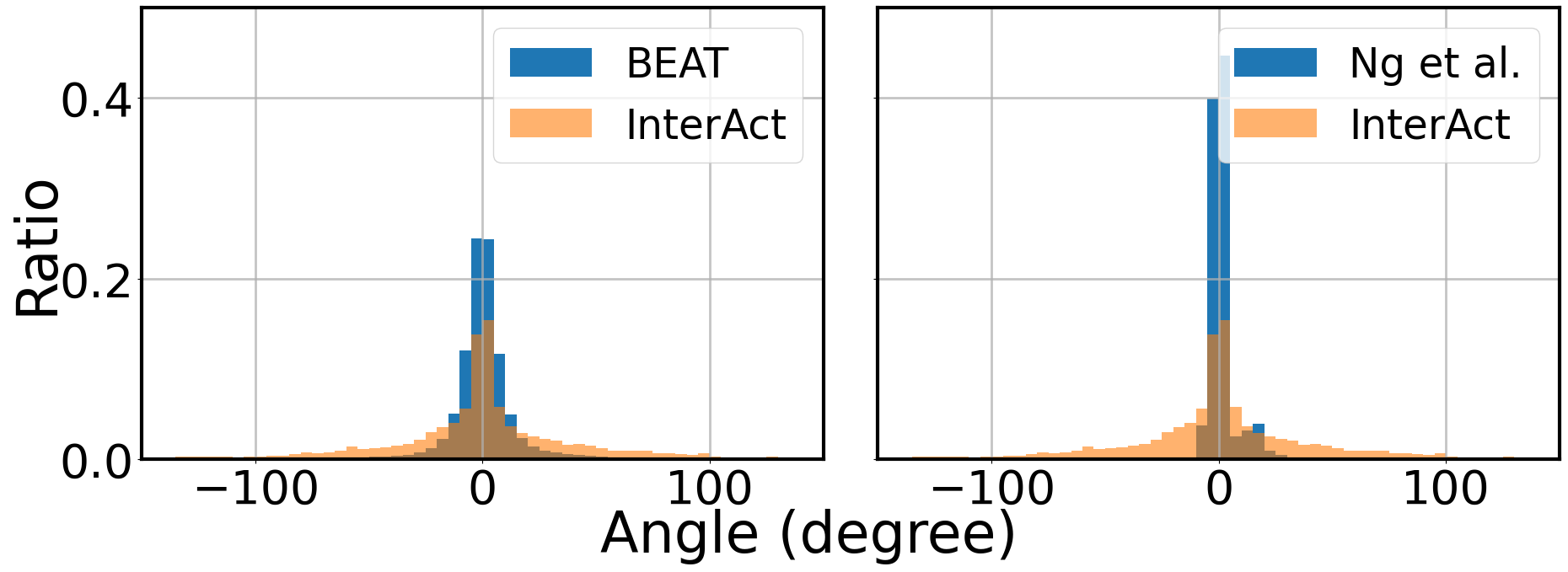}       
        \caption{Histograms of rotations.}
        \label{fig:subfig_c}
    \end{subfigure}\hfill
    \begin{subfigure}[b]{0.49\linewidth}
        \includegraphics[height=0.36\linewidth]{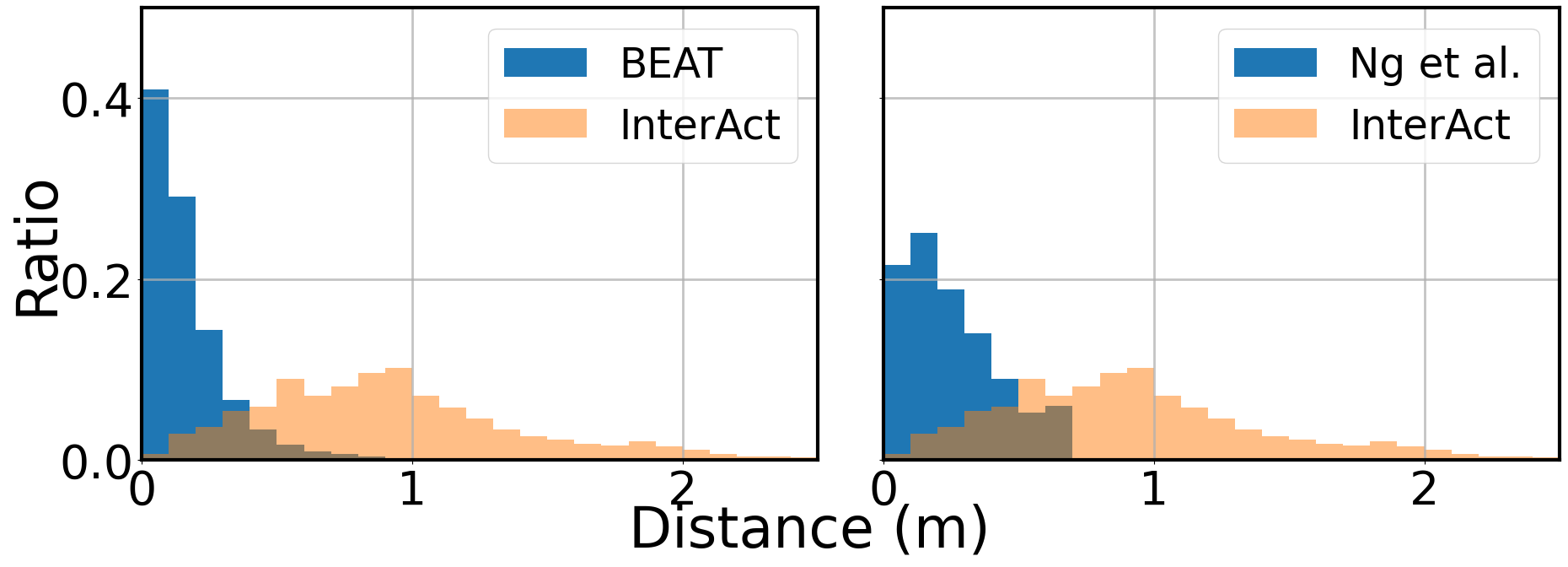}      
        \caption{Histograms of distances.}
        \label{fig:subfig_d}
    \end{subfigure}

    \caption{Statistical comparisons between InterAct and previous datasets w.r.t. individual motions (upper part) and relative body orientation and distance between two actors (lower part). Body motions in InterAct exhibit a bigger diversity and are also more dynamic. Note that in each session, only one actor is captured in BEAT and in \cite{ng2024audio2photoreal}, so analysis in the lower part is not applicable on them.}
    \label{fig:stat_body_single}
\end{figure}

We report the comparison results in Fig. \ref{fig:stat_body_single}.
Relative distance and body orientations are important indicators when investigating two-person interactions. Here we conduct statistical analysis on InterAct, BEAT and the one in \cite{ng2024audio2photoreal}. 
Firstly we treat each person's motions individually. For each sequence, we compute the relative distance and orientation of the root joint for the first frame and the remaining frames. The results are reported in the upper part of Fig. \ref{fig:stat_body_single}. We similarly compute the relative distance and orientation of two actors for all sequences. From the statistical results, we can easily see that individual motions in InterAct enjoy much bigger diversity than those in BEAT and the dataset in \cite{ng2024audio2photoreal}. The comparisons highlight one key difference between InterAct and the others: InterAct contains both conversational gestures and dynamic movements which are relatively large-scale and long-term, while BEAT and the one in \cite{ng2024audio2photoreal} focus on gestures and small-scale movements.

\subsubsection{Entropy and Variance of Animations} 
\begin{figure}
    \begin{subfigure}[c]{0.49\linewidth}
        \includegraphics[width=1\linewidth]{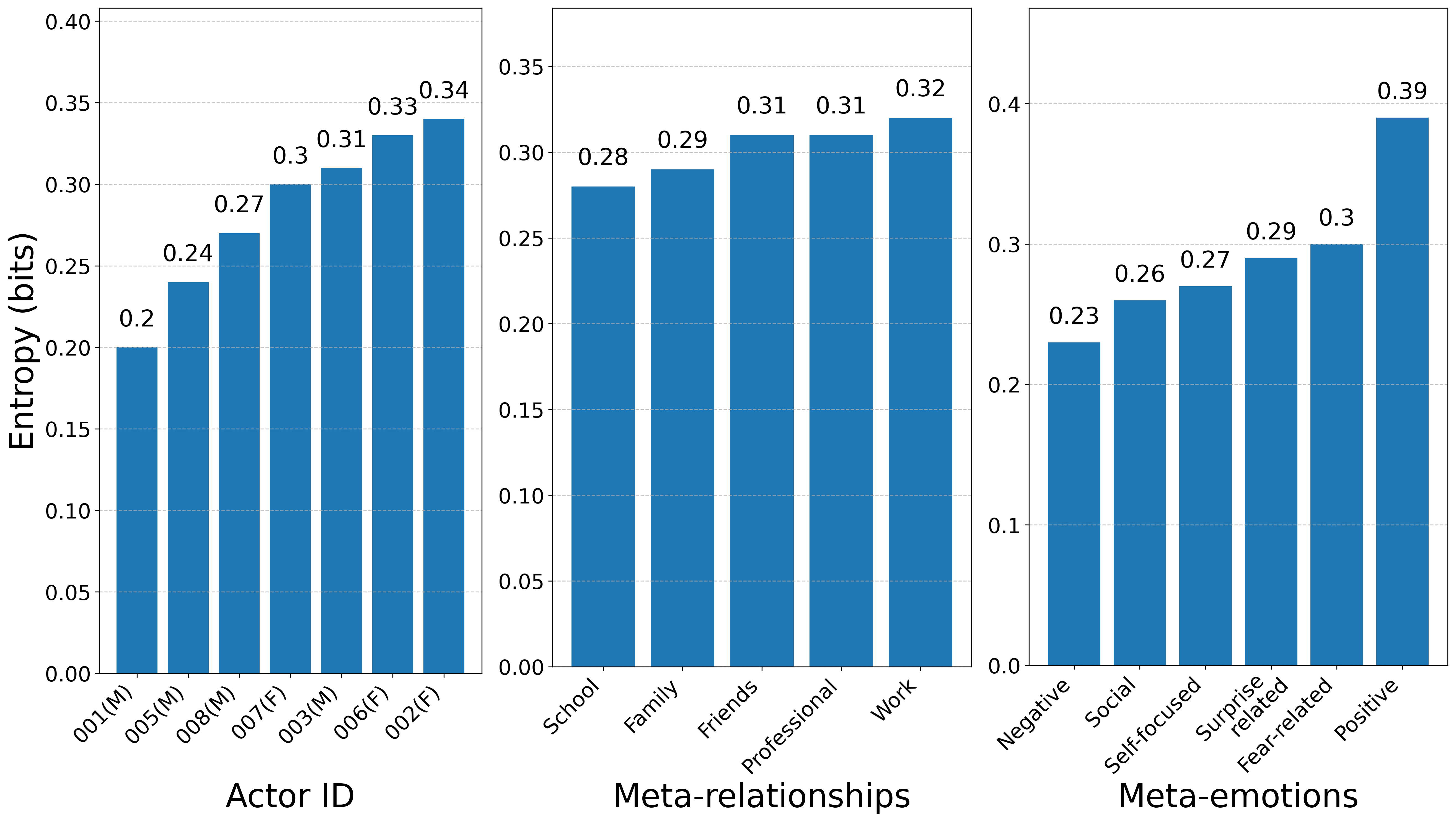}
        \caption{Entropy of body motions w.r.t. different actors, meta-relationships, and meta-emotions. M = Male.}
        \label{fig:entropy_body}
    \end{subfigure}\hfill
    \begin{subfigure}[c]{0.49\linewidth}
        \includegraphics[width=1\linewidth]{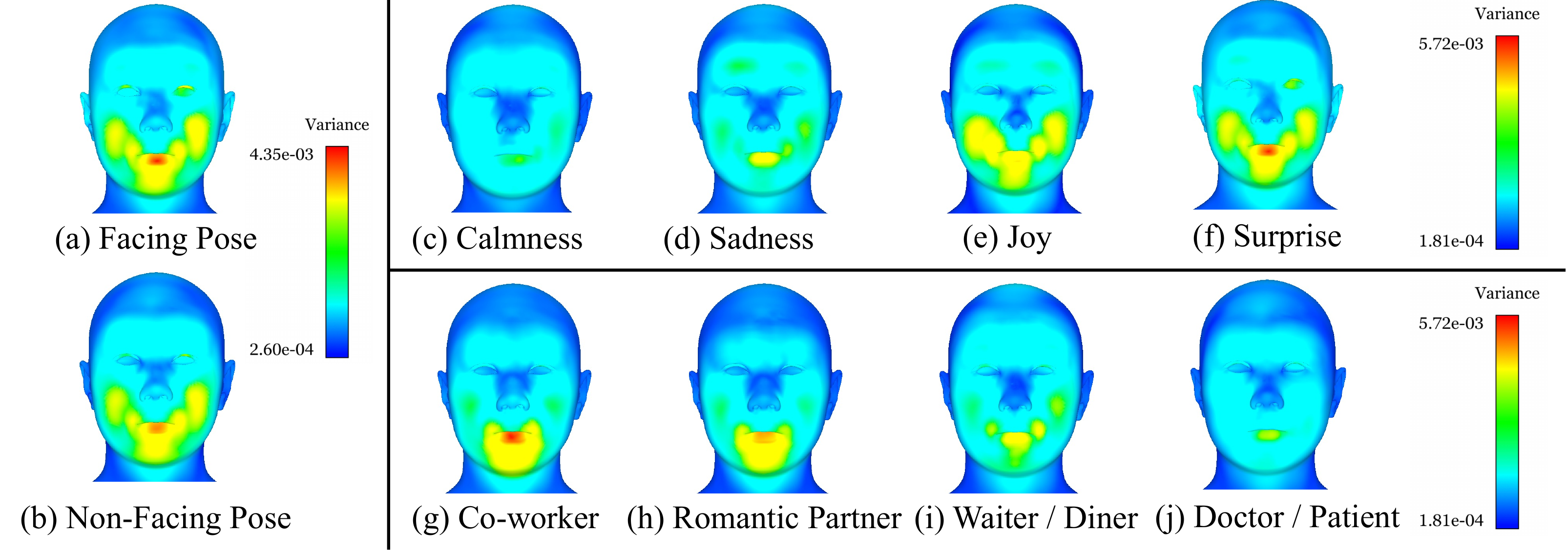}
        \caption{Variance of different parts of the face w.r.t. different relationships, emotions, and facing poses.}
        \label{fig:face_variance}
    \end{subfigure}
    \caption{Left: We calculate entropy values for body motions of different actors, meta-relationships, and meta-emotions. Higher entropy indicates more varied and unpredictable motions, and generally more information contained. We find that our female actors (indicated by "F"), professional and work settings, and emotions pertaining to surprise, fear, or positivity show higher entropy. Right: We visualize the variance of facial mesh vertices in relation to different relationships, emotions, and whether or not actors are facing each other. We find that emotions such as joy, surprise, and work relationships such as co-worker shows higher variance, a result that agrees with body entropy findings.}
    \label{fig:relative_ratio}
\end{figure}

We also conduct a statistical analysis on the body and facial animations in our dataset to demonstrate the expression and diversity. There are two main aspects:

\paragraph{Body Entropy} Shannon entropy values for body motions of different actors, meta-relationships, and meta-emotions are calculated by first computing the top 20 joints with the highest average angular velocity, then discretizing the instantaneous angular velocities of these joints into 100 bins for normalization to a probability distribution. See Fig. \ref{fig:entropy_body} for the results. Our results show that our female actors, work and professional settings, as well as positive, fear and surprise-related emotions produce motions with the highest entropy. This indicates a more varied and dynamic performance, which potentially contains more information. It can also be used as a general guide to show which motion categories may be harder to learn compared to others.

\paragraph{Face Variance} Face Variance between Different Facing Directions, Emotions, and Relationships indicates a slightly higher variance in the lip area for the facing pose, demonstrating greater lip movement when actors are facing each other. The metrics are calculated by first aggregating the position delta of each vertex of the facial mesh between the current frame and a static face expression of the same actor, then determining the variance across all frames of the same category in question.
We visualize it in the right part of Fig. \ref{fig:relative_ratio}. The variance of the lip area for surprise is the highest among all emotions, indicating a tendency for actors to open their mouths widely when provoked; Co-workers also show larger variations around the lip area. These results agree with our findings with respect to body entropy.

\subsubsection{Self-contacts and Cross-contacts} While investigating contacts is not a primary focus of our work, it nonetheless provides an interesting perspective with which our two-person data can be analyzed. To that end, we computed the duration and rate of hand-head, hand-shoulder, hand-elbow, and hand-hand contacts between the actor and their own body parts, and their partner's body parts respectively. Tab. \ref{tab:self-contacts} and Tab. \ref{tab:cross-contacts} in our supplementary show the full self-contacts and cross-contacts matrices respectively. We note that there are no significant differences of hand contact patterns between male and female actors. Furthermore, we see significantly more hand-hand self-contacts in scenarios with friends (22.8 min/hr), schoolmates (17.6 min/hr), and social emotions (11.4 min/hr). Such contacts are likely in the form of steepling, indicating engagement and confidence, or a relaxed clasp, suggesting composure and attentiveness within a comfortable setting. We also find hand-elbow self-contacts and cross-contacts to be more frequent during fear-related scenarios, highlighting the defensive nature of such a gesture.

\subsubsection{Qualitative Analysis} Many instances of interesting dynamic two-person interactions are captured in our new dataset. We show a gallery of such interactions in Fig. \ref{fig:teaser}. Interesting subconscious phenomena are captured, such as a person shifting their body away from another person to protect themselves in anticipation of being hit. We also capture reactive motions, such as acting shocked when a person scares them, completing a high-five, being coached on how to perform certain motions, posing as the other person takes a picture for them, and more. Emotional movements and expressions are also exhibited categorically, such as a person jumping in the air when experiencing great joy, crossing their arms and eyebrows when feeling angry, or hugging intimately when feeling romantic. Furthermore, we also capture collaborative interactions such as two person collectively trying to work on a stationary large object while using smaller imaginary objects as tools, such as using a screwdriver to affix screws to a table leg. Compared to BEAT where only one actor orates speeches with body gestures and movements, or in \cite{ng2024audio2photoreal} where two actors converse standing still, our dataset shows a much wider variety of daily scenarios which require the interaction and collaboration of two people.

\subsubsection{Hands and Facial Quality} We provide a remark on the motion quality of the dataset's hands and facial data. For hands, we note that a minor portion of capture performances (less than 5\% of data) had been affected by anomalies, such as label swaps on one hand, index finger extensions beyond anatomical limits, and flickering. We believe the label swaps and finger extension errors stem from partially worn-out finger markers being used during capture, and flickering from the occasional close proximity of an actor to a particular MoCap camera. However, the majority of our hands data remain anatomically accurate and usable. For the face, the conversion from facial meshes to ARKit blendshape parameters introduces tiny quantitative errors, primarily around the lip area. However, side-by-side comparisons have shown the differences to be practically imperceivable, and do not affect important facial quality markers such as lip closure for plosives. Please see our supplementary video for examples of hand and face motion.

\section{Conversational Facial and Body Motion Generation Baseline }

In a dyadic interaction scenario, given the speech inputs $a_A$ and $a_B$ from two individuals, our task is to generate their corresponding \textbf{facial expression animations} ($f_A$, $f_B$) and \textbf{full-body motions} ($m_A$, $m_B$). Formally, this can be represented as:
\begin{equation}
\{f_A, f_B, m_A, m_B\} = \mathcal{G}(a_A, a_B, C),
\label{eq:generation}
\end{equation}
where $a_A$ and $a_B$ are the raw speech signals of character A and B, respectively. $C$ denotes additional \textbf{contextual signals} and  $\mathcal{G}$ is the generation model that jointly synthesizes \textbf{synchronized} and \textbf{interaction-aware} non-verbal behaviors for both participants. Exponential map \cite{grassia1998practical} is adopted as the motion representation~\cite{alexanderson2020style, alexanderson2023listen}.

\begin{figure}
    \centering
    \includegraphics[width=1.0\linewidth]%
    {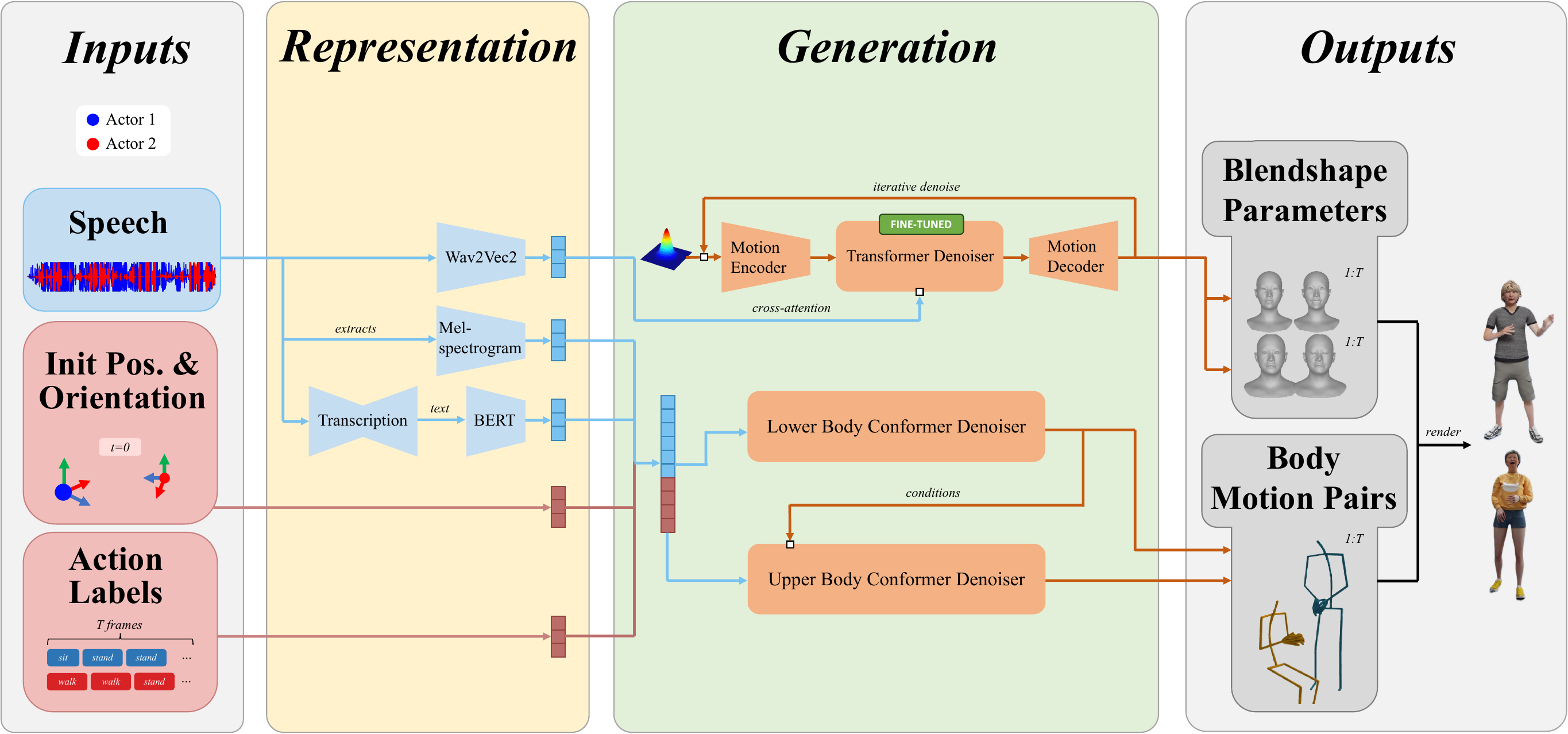}    
    \caption{Pipeline of our system. Given raw audio signals of two people, BERT feature, relative orientation and position, and action labels as conditions, our method can generate realistic and diverse facial and body animation by cooperating two diffusion models. The result of our algorithm is the simultaneous prediction of facial blendshape parameters and comprehensive global body movements for both subjects. Note the hierarchical generation of body motions using two networks, and the fine-tuned denoiser in our facial pipeline.}
    \label{fig:pipeline}
\end{figure}

\subsection{Facial Motion Synthesis}

We propose a model to regress the facial expressions of an actor driven by dialogue, as well as a body feature. We note that a person's facial expressions and eye movement can differ substantially depending on whether they are currently facing the other person, i.e., the direction of their gaze, the context of the dialogue, as well as the spoken content and facial expression of the other person. Using a fine-tuning method and a novel inference technique, we also improve the lip accuracy and sidestep the issue where Transformer-based models fail to close the mouth when pronouncing phoenemes like b, p after training on emotional datasets.

The model for regressing a person's facial animations is built on top of the DiffSpeaker~\cite{ma2024diffspeaker}, where the auto-regressive FaceFormer \cite{fan2022faceformer} architecture with alignment and attention bias used for frame-by-frame animation generation is adapted as a diffusion model. We adapt the model to support additional inputs.

Specifically, for input person $A$ and conditioned person $B$, we aim to produce the facial motion $f_{A}^{1:T}$ of time duration $T$ given the synchronized two-person dialogue audio $a_A^{1:T}$ and $a_B^{1:T}$, the one-hot embeddings of the speakers $s_A$ and $s_B$, as well as a one-hot embedding $p \in \mathbb{R}^2$ denoting whether the two persons are facing each other. The entire network $\textbf{F}$ can be formulated as: 

\begin{equation}
    \hat{f}_{A} = \textbf{F}(a_A, a_B, s_A, s_B, p).
\end{equation}

During training, the person's ARKit sequence $f_{A}^{1:T}$ is encoded into 512-d, giving the motion encoding $x_{A}$. For audio, $a_A^{1:T}$ and $a_B^{1:T}$ are first seperately encoded using the pretrained Wav2Vec2~\cite{baevski2020wav2vec} to produce vectors $w_A^{1:T}$ and $w_B^{1:T}$. Afterwards, the vectors are concatenated and passed through a linear layer to produce the final audio representation $e_a$. Similarly for conditions, for the speaking style $s_A$ and $s_B$, the one-hot embedding $p$, and the diffusion step information $n$, they are passed through their respective condition encoders to produce $e_s$, $e_p$, and $e_n$. The diffusion denoiser $\textbf{D}$ can therefore be formulated for training as:
\begin{equation}
    \hat{x}_{A}^0 = \textbf{D}(x_{A}^n, e_a, e_p, e_n)
\end{equation}
where $x^n$ denotes the latent facial animation $x$ added with Gaussian noise, with $x^0$ denoting the latent animation without noise added, and $x^N$ denoting pure noise. The predicted latent $\hat{x}_{A}^0$ is then decoded back into ARKit parameters to give $\hat{f}_{A}$.

For the cross-attention mechanism in the denoiser Transformer architecture, we extend the \textit{Biased Conditional Attention} in Diffspeaker which utilizes Prefix Tuning as proposed by \cite{pfeiffer2020AdapterHub} to include the pose encoding $e_p$ as an extra condition. The attention score ($\textbf{S}$) formula is hence modified as:
\begin{equation}
    \textbf{S} = \operatorname{Softmax}(Q{[e_s, e_p, e_n, K]^T} + B),
\end{equation}
where $Q$, $K$, $B$ denote the query, key, and bias terms of the attention score respectively.

After training a base model, we fine-tune the model with our small, lip-accurate dataset, using loss defined solely on the lips. At inference time, we use the weights of the base model while only swapping the weights of the denoiser with the fine-tuned model, keeping the audio encoder and other projection layers untouched. This technique improves lip accuracy while preserving the lip shape of individual actors, counteracting the mode collapse inherent with simple fine-tuning, where the lip shape of all actors eventually morphs to that of the actor in the fine-tuning dataset. See our supplementary for example.

\subsection{Body Motion Synthesis}

We aim at audio-driven interactive motion estimation between two people. 
Following the success ~\cite{alexanderson2023listen} by using a diffusion model on the speech2motion task, we extend it to jointly estimate the full-body motions of two people from their speech signals. 
Each actor's audio will be processed separately, and then concatenated as network inputs. We follow~\cite{pang2023bodyformer} to choose the mel-spectrogram as the proxy for low-level features, and the BERT~\cite{devlin2018bert} features calculated from text transcripts for semantics.
For better modeling and generating meaningful human motions, we compute the action mode in a heuristic way and assign a label $C$ among \{sit, walk, stand\} to each frame.
In the inference stage, we can assign different action modes to get diverse results.
Formally, the motion synthesis is defined as follows:
\begin{equation}
    [m_A, m_B] = \mathcal{G}_M(\text{Mel}(a_A, a_B), \text{B}(T_A, T_B), C),
\end{equation}
where $\mathcal{G}_M$ is the motion generator, $\text{Mel}$ is the mel-spectrogram extractor \cite{librosa}, $B$ is the BERT feature extractor, $T_A$ and $T_B$ are transcripts accompanied with $a_A, a_B$, and $C\in[0,1]^3$ is the action label.
In practice, the model is trained with a diffusion process. Specifically, noised motion frames $m_A, m_B$ are fed into the generator $\mathcal{G}_M$, and the difference between the denoised output and the ground-truth motion is treated as the training loss to train the diffusion models to learn the denoising process.

\paragraph{Hierarchical Interactive Motion Estimation}
Though conceptually simple and easy to implement, this naive approach does not yield good results in practice. This can be partially due to the fact that the output dimension is quite high (316-d in our case vs 70-d in the original method) when two people with detailed hands are jointly considered. In addition, the statistical attributes of different joints might be pretty different during various kinds of interactions.
Hence, we divide joints into the lower and upper body. We propose to first regress lower-body joints from the control signal, and then regress the upper-body ones conditioned on both the control signal and the regressed lower-body joints. We empirically find this hierarchical mechanism works better.

\section{Experiments}
We train and validate audio-driven interactive human motions and facial expressions on the new dataset, and conduct the quantitative and qualitative evaluation. Once trained, our algorithm can directly yield global body motion and detailed facial animations for both actors. The initial relative configuration (orientation and position) between two actors is assumed to be given to make this heavily ill-posed problem better constrained.

\subsection{Audio-driven Facial Expression Generation}

We provide a quantitative evaluation of the face motion generated by our system.

\paragraph{Metrics}
Facial expression accuracy and fidelity can be measured by considering the lip synchronization and upper face variations separately. Lip Vertex Error (LVE) is measured as the averaged maximum $l_2$ error among all lip vertices of the facial mesh across all frames. This is a common metric used in ~\cite{richard2021meshtalk, fan2022faceformer, ma2024diffspeaker}.
Face Dynamics Deviation (FDD), proposed by \cite{xing2023codetalker}, aims to quantify the variation of upper facial dynamics by means of comparing deviation over time, by calculating the temporal standard deviation of the $l_2$ element-wise norm. Tab. \ref{tab:quantitative_face} shows our result on the test set. Our fine-tuned model achieves the highest lip accuracy while both models outperform FaceFormer in terms of upper facial dynamics.

\begin{table*}[t]
    \centering
    \begin{minipage}[c]{0.49\textwidth} %
        \captionof{table}{Numerical comparison of different methods. $\text{LDA}_1$ is LDA trained and tested on individual persons; $\text{LDA}_2$ is LDA adapted to regress all joints of two people; Ours means our hierarchical method. Ours performs the best.}
        \resizebox{\textwidth}{!}{
        \setlength{\tabcolsep}{4pt}
        \centering
        \begin{tabular}{cccccc}
            \toprule
           Method & $\text{FID}_g\downarrow$ & $\text{FID}_k\downarrow$ & $\text{
            Div}_g\uparrow$  & $\text{Div}_k\uparrow$ & $\text{Div}_{spl}\uparrow$  \\
            \midrule
            EMAGE & 11.01 & 17.26 & 14.80 & 1.48 & 14.79\\
            Ng et. al. & 17.21 & 28.99 & 10.32 & 0.29 & 12.43 \\    
            $\text{LDA}_1$ & 11.78 & 20.11 & 21.45 & 1.53 & 20.83\\        
            $\text{LDA}_2$ & 12.22 & 20.80 & \textbf{23.52} & \textbf{1.56} & 21.50\\
            \rowcolor[gray]{0.80}
            $\text{Ours}$ & \textbf{6.84} & \textbf{11.18} & 23.49 & 1.52 & \textbf{23.27} \\ 
            \bottomrule
        \end{tabular}
        }
        \label{tab:quan_comp}
    \end{minipage}%
    \hfill %
    \begin{minipage}[c]{0.49\textwidth} %
        \captionof{table}{Result of user study for generated both motions. Users find our method to be much better than LDA$_{\text{2}}$. "M. Better" means "Much Better".}
        \resizebox{\textwidth}{!}{
        \setlength{\tabcolsep}{2pt}
        \centering
        \begin{tabular}{cccccc}
            \toprule
            Comparison & M. Better & Better & Same  & Worse & M. Worse  \\
            \midrule
            $\text{Ours vs LDA}_2$ & \textbf{61.18\%} & 13.82\% & 12.50\% & 11.84\% & 0.66\% \\    
            \bottomrule
        \end{tabular}
        }
        \label{tab:body_user_study}
    \end{minipage}
\end{table*}

\begin{figure}[t]
    \centering
    \begin{minipage}[c]{0.48\linewidth} %
        \centering
        \includegraphics[width=\linewidth, trim={0cm 0 0 0cm}, clip]{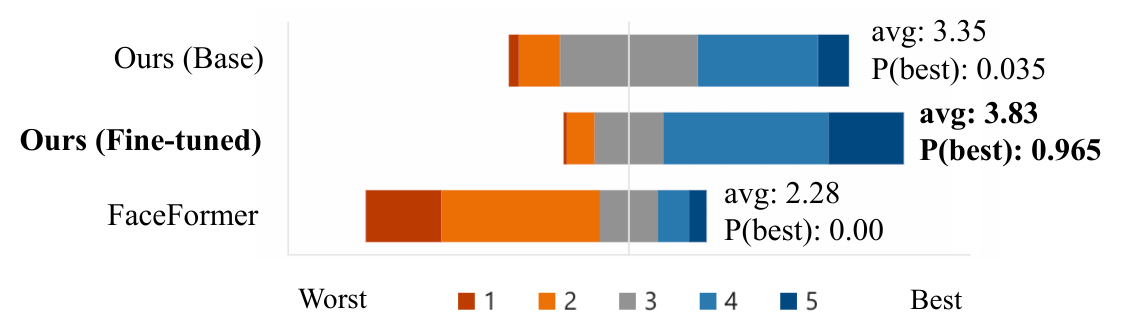}
        \caption{Result of face user study. Users are asked to rate each method's realism from worst (1) to best (5). P(best) is then calculated using Dirichlet sampling.}
        \label{fig:face_user_study}
    \end{minipage}\hfill %
    \begin{minipage}[c]{0.48\linewidth} %
        \captionof{table}{Numerical comparison of different face methods using ARKit format, except for FaceFormer (Vertex).}
        \centering
        \setlength{\tabcolsep}{6pt}
        \resizebox{\linewidth}{!}{ %
        \centering
        \begin{tabular}{ccc}
            \toprule
            Method & LVE $\downarrow$ & FDD $\downarrow$ \\
            \midrule
            FaceFormer (Vertex)     & $3.5927\times10^{-5}$   & $9.6407\times10^{-5}$ \\
            FaceFormer (Blendshape) & $5.1961\times10^{-5}$   & $7.4392\times10^{-5}$ \\
            \rowcolor[gray]{0.80}
            Ours (Base)             & $3.6058\times10^{-5}$   & $\mathbf{5.2460\times10^{-5}}$ \\
            \rowcolor[gray]{0.80}
            Ours (Fine-tuned)       & $\mathbf{2.8757\times10^{-5}}$   &$5.6982\times10^{-5}$ \\
            \bottomrule
        \end{tabular}
        }
        \label{tab:quantitative_face}
    \end{minipage}
\end{figure}

\paragraph{User Study}
We conduct user studies between the results generated by our base and fine-tuned methods and blendshape FaceFormer. Altogether 20 users each review 5 random results for every method side-by-side and are asked to rate each method's realism based on emotion portrayed, lip accuracy, and semantic congruence. Fig. \ref{fig:face_user_study} shows our results. One sample of our generated results is shown in Fig. \ref{fig:compare_955}. Please check our supplementary materials for more results.

\subsection{Audio-driven Body Motion Generation}
We conduct an extensive evaluation of the full-body motion generation of our system trained on InterAct.  We first provide the metrics of evaluation and then the comparison of our system with the SOTAs.  

\paragraph{Metrics and Baseline Methods} 
We use the same set of metrics proposed in \cite{ng2024audio2photoreal} to measure the realism and diversity of the generated motions. We re-train LDA in \cite{alexanderson2023listen}, EMAGE in \cite{liu2023emage} and the method in \cite{ng2024audio2photoreal} on InterAct and report the numerical results in Tab. \ref{tab:quan_comp}. We notice that both EMAGE and the one in \cite{ng2024audio2photoreal} do not perform well on InterAct. This can be attributed to two reasons: 1) motions in InterAct are more diverse and thus more challenging; 2) actors in InterAct change body orientation and location heavily, while EMAGE and the method in \cite{ng2024audio2photoreal} focus on conversational gestures where the body movement is minor. The relative-to-last-frame root representation used in LDA and our method can better handle large body movement since only difference between consecutive frames is modelled.

\begin{figure}[t]
    \centering
    \begin{subfigure}[b]{1.\linewidth}
        \includegraphics[width=1.0\linewidth]{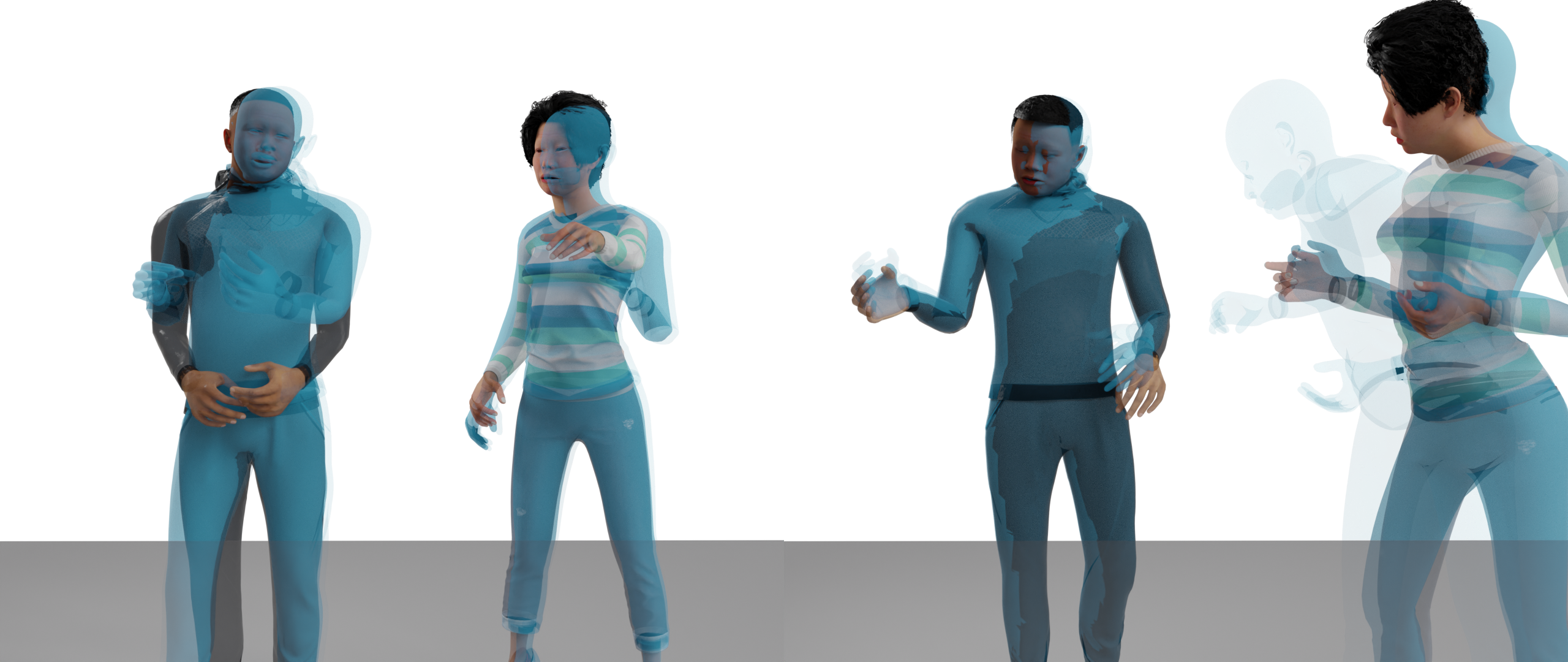}
    \end{subfigure}            
    
    \begin{subfigure}[b]{1.\linewidth}
        \centering
        \includegraphics[width=0.3265\linewidth]{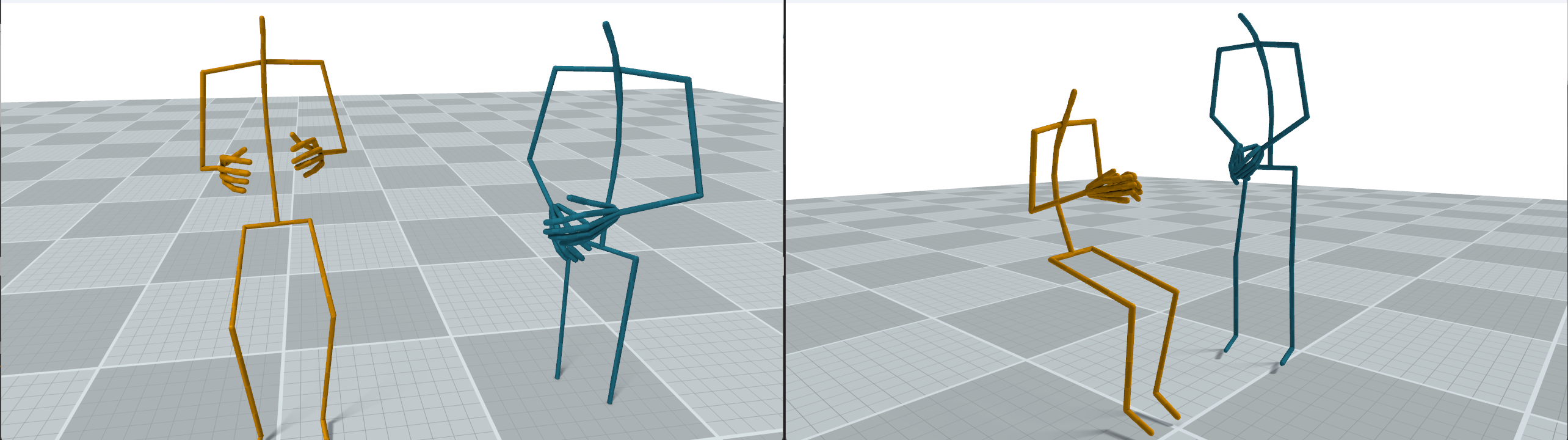}    
        \includegraphics[width=0.3265\linewidth]{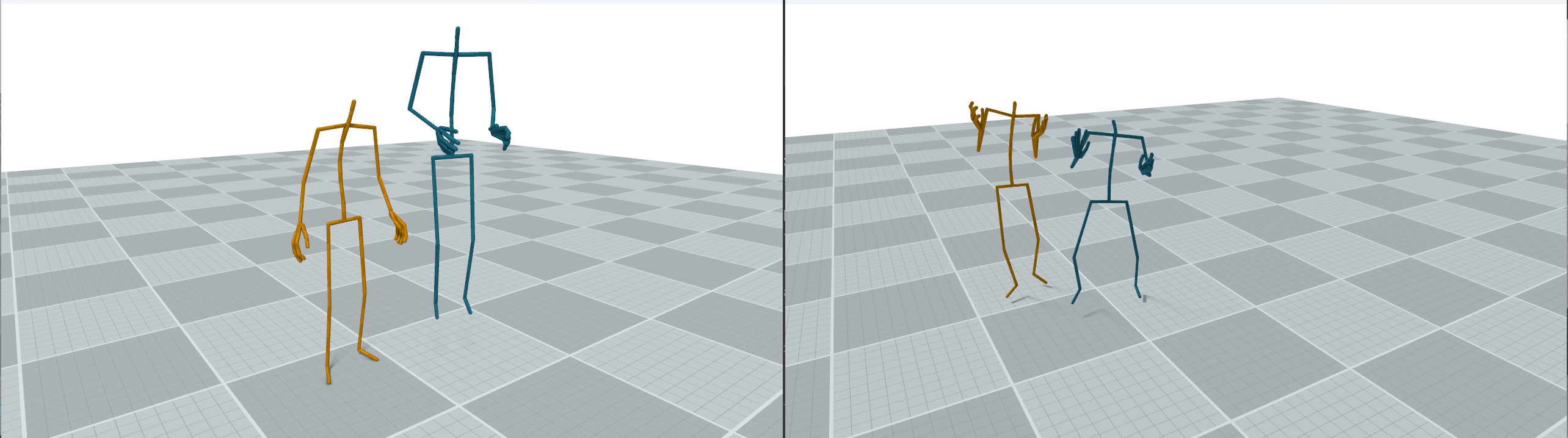}   
        \includegraphics[width=0.3265\linewidth]{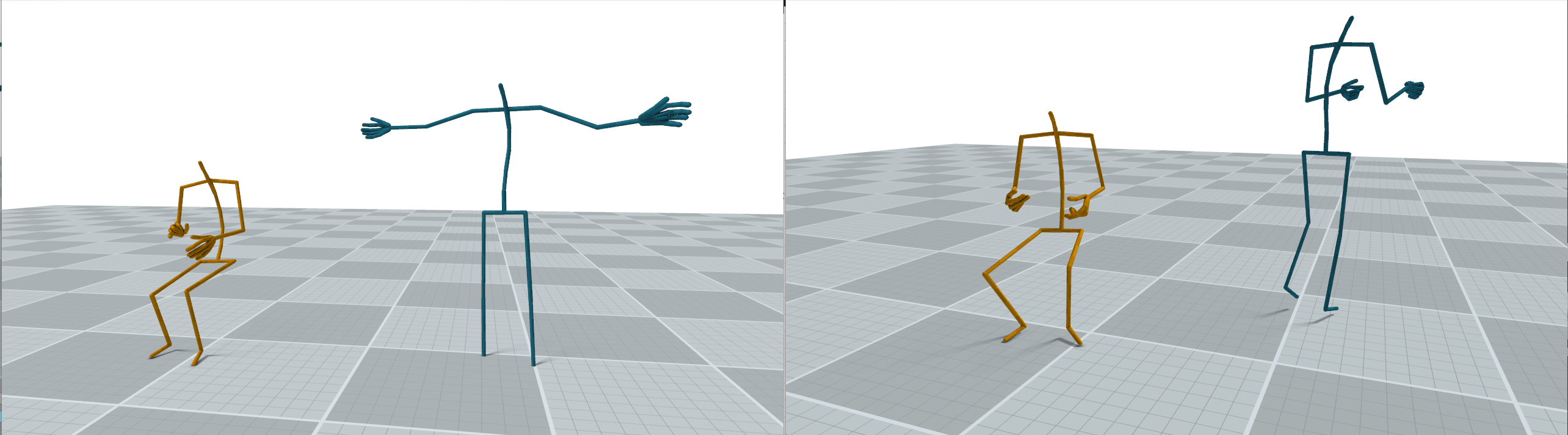}         
    \end{subfigure}            

    \caption{Top: Sample of our results where two actors are having a conversation. Bottom: Skeleton comparison between the predictions of our method with the ground-truth motions. For each pair of images, the left one is our predicted result, and the right one is the ground truth. Conditioning on the same audio input, we generate plausible full-body motions different from GT. Left two pairs are from the waiter / diner relationship, and the rightmost pair is of the co-worker relationship. The scenario goes like this: \textit{the older co-worker is eating a disgustingly smelly lunch at his desk, and the unfazed reactions of those around him shocks and disgusts the new hire}.}
    \label{fig:compare_955}
\end{figure}

\paragraph{Qualitative Comparison}
To assess the visual realism and precision of our technique, we rendered both the predicted motions from our method and the ground truth motions corresponding to the same speech, as illustrated in Fig. \ref{fig:compare_955}.
Thanks to our extensive and varied dataset, coupled with a robust baseline system, our method is able to generate motions that are not only convincing but also exhibit diversity when given the same input.
Our approach also offers the flexibility to tailor the generated motions by applying different contextual conditions, such as the nature of the relationship, action labels, and emotional states.
Please refer to our supplementary video for more details. 

\paragraph{User Study}  We conduct user studies between results generated by our method and $\text{LDA}_2$.
EMAGE and the one in \cite{ng2024audio2photoreal} are excluded since theirs are proposed for a different task from ours.
We render the results of different methods and show them side by side, and then ask the anonymous evaluators to rate the quality of the videos. Overall 38 people participated in the user study. The result is shown in Tab. \ref{tab:body_user_study}. %

\if
Following prior works ~\cite{alexanderson2023listen, ng2024audio2photoreal}, we employ a suite of metrics that capture both the realism and diversity of generated motions. 
1. The Fréchet Inception Distance (\textbf{$FID$}) is a widely accepted metric in generative tasks, 
quantifying the distributional distance between ground truth and generated data, with various features extracted to represent different aspects of the generated motion. 
The $FID_{g}$ focuses on "geometric" realism by measuring this distance for static poses of two people, adjusted for orientation to face the positive x-axis. 
The $FID_{k}$ metric extends the evaluation to "kinetic" realism by collaborating with a pre-trained feature extractor. This extractor encodes the pose sequences of a single individual into a latent representation, facilitating the measurement of motion quality in a dynamic context.
For evaluating "two-person relationship" realism, $FID_{r}$ calculates the joint distances map between two individuals. 
2. We also assess the diversity $DIV$ by computing the average pairwise distance of generated samples. 
3. Lastly, the $Foot.Slid$ metric gauges physical plausibility by quantifying the extent of foot sliding or skidding, with lower values indicating more realistic foot behavior.

In Tab.~\ref{tab:quantitative_body}, we conducted an evaluation of body motion generation. 
In comparison to the baseline LDA model, which also incorporates audio input, our method exhibited lower performance for the single person's motion, in terms of the dynamic and physical realism.
This discrepancy stems from a more complex conditional space in  task setting, that our model is designed to generate motion that is cognizant of the presence of another individual.
Despite this, our method demonstrated a superior capability in capturing the nuances of two-person interactions. 
It produces more convincing static poses and maintains better relative joint distances. 
Furthermore, the diversity of interactions between two individuals is enhanced through the incorporation of relative offsets as an additional input, enriching the overall interaction modeling.
\fi

\section{Ethical Considerations}
Prior to conducting research, we have undertaken several important ethical considerations. First, we have obtained informed consent from all participants involved in the study. Participants are fully aware of the nature of the research, the types of data being collected (including audio, body motions, and facial expressions), and how their data will be utilized. In the final dataset, all data is anonymized to prevent the identification of individuals involved in the interactions. We recognize the potential ethical concerns regarding AI generated content and will continuously work with the broader research community to ensure our data handling and release procedures align with best practices.

\section{Limitations}

We note several limitations pertaining to the capture and content of the dataset, as well as the baseline generation method.

\subsection{Dataset} For the dataset, the data collected is inherently limited by \textbf{Capture Setup Constraints}. The use of head-mounted cameras prevents close facial interactions such as kissing or whispering in the ear. Moreover, having two people in the capture area increases the potential for marker occlusion, negatively affecting the accuracy. The \textbf{Nature and Scope of Performances}, such as the acting nature of our data as opposed to capturing reality, provides an alternative view to many scenarios but may be regarded as less realistic. Furthermore, the lack of props and capturing on a flat ground limits the applicability of this dataset for human-object interaction tasks or scenarios with different environmental attributes. Finally, while we strive to improve the diversity of our dataset, we note that our coverage of \textbf{Actor Demographics} is imperfect and limited by our sample size of 7 actors. For example, the dataset includes scenarios in which adult actors portray characters of different ages such as children. Although professionally executed, these portrayals may not fully capture the unique motor patterns, energy levels, or nuanced social dynamics inherent to genuine child interactions.

\subsection{Generation Baseline} For the generation baseline method, limitations concern mainly the \textbf{Reliance on Speech and Simple Contextual Cues}. Our current baseline heavily leverages speech and basic action labels to drive motion, which, while demonstrating potential, means that speech only roughly indicates the full complexity of dyadic behaviors. Finer-grained control and more nuanced animations necessitate more sophisticated conditioning on richer contextual signals. The \textbf{Accuracy of Fine-Grained Motions and Physical Plausibility} also remains to be improved. Specifically, the generation of precise hand and finger poses remains a challenge, and when generated motions are transferred to 3D characters, issues such as self-penetrations or inter-character collisions can arise, showing the need for enhanced physical realism. Lastly, the \textbf{Generalization and Controllability} of the baseline model remains an under-explored area in our work. The model's ability to adapt to significantly different interaction styles or unseen actors may be limited, and enhancing control beyond current inputs is crucial for broader applicability.

\section{Conclusions and Future Works}
In this paper, we target capturing and modeling the ubiquitous interactive activities between two people in various daily scenarios
by collecting novel multi-modal interaction data and designing a hierarchical diffusion model.

The presented InterAct dataset, a key contribution of this work, offers significant utility for future works across several research and development domains. For \textbf{Human Dyad Research}, our dataset captures human speech, expression, and motion in diverse daily life settings under varying relational and emotional contexts. This provides rich data for researchers studying how interpersonal connections and mood influence non-verbal communication strategies, proxemics, and collaboration. For the task of \textbf{Full-Body Synthesis between Two People}, InterAct facilitates advancements in synthesizing dyadic interactions, as demonstrated by our baseline audio-driven motion estimation model. Future researchers can leverage its unique accuracy and multi-modality to integrate with other datasets, develop architectural improvements, and benchmark models for nuanced two-person motion generation. For \textbf{Socially-Aware Motion Generation} in robotics and VR, InterAct offers real-life, physically-informed data crucial for developing human-centric, socially-aware robots and digital agents, with applications spanning numerous industries. Lastly, for \textbf{Advancing Challenging CV Tasks}, InterAct facilitates research in multiple ill-posed computer vision problems, such as dual-subject human mesh recovery from monocular input, by providing precise MoCap ground-truth data that directly contributes to the training and evaluating of robust models capable of understanding complex multi-person scenes.

\clearpage

\bibliographystyle{ACM-Reference-Format}
\bibliography{main}

\clearpage
\setcounter{page}{1}
\appendix

\section*{Supplementary}

In this supplementary material, we provide additional information about the InterAct dataset, the network architecture, training details, and additional results.

\section{Dataset Details}

\subsection{Character Setup and Scenario Design}

We design a one-sentence character setup and a scenario description. Tab. \ref{tab:scenarios} includes multiple exemplary scenarios captured in our dataset. The full scenarios list is included in the supplementary PDFs.

\begin{table}[b]
    \caption{Excerpt of scenarios in our dataset}
    \centering
    \begin{tabular}{p{2cm}p{2cm}p{4.5cm}p{4.5cm}}
        \toprule
       Relation & Emotion & Character Setup & Scenario Design \\
        \midrule
        Siblings & Happiness & \small An older brother, who is 20 years old, and his little sister, who is 16 years old & \small They both unexpectedly receive acceptance letters from their dream schools \\

        \midrule
        
        Coworker & Romance & \small One project manager in her 30s and a software engineer in his 40s. They have known each other for several years & \small In one late-night working session at the empty office, they confess their feelings for each other \\

        \midrule
        
        Principal / Student & Awkwardness & \small A high school principal, in their late 40s, and a 16-year-old student who is generally reserved. & \small During lunch break, both have discovered they’re the only ones in the cafeteria and, upon meeting eyes, got drawn into an unintentional conversation about the student's recent performance in school. \\

        \midrule

       Churchmates & Interest & \small Two churchmates, an elderly choir singer and a young, recently-turned atheist. & \small They are engaged in a deep but respectful conversation about faith and belief during a church gathering. \\

        \midrule
       
        Doctor / Patient & Relief & \small A doctor in his mid-40s, who is known for his professionalism, and a patient, a woman in her early 60s, who has been anxious about her medical diagnosis. & \small The doctor delivers the medical test results, revealing that the patient does not have a terminal illness as feared, but a treatable condition. \\
        \midrule
        Waiter (waitress) / Diner & Anger & \small A middle-aged waitress who has been having a tough day and a young, impatient customer who has never worked in the service industry. & \small The waitress has accidentally spilt a small amount of drink on the customer's jacket, and the customer is overreacting and being quite harsh to the waitress. \\
        \midrule
       Assisted Living Facility Neighbors & Sadness & \small Two elderly characters, both residents of an assisted living facility and have become close friends over time. & \small One of them shares the news that their grandchild has just stopped visiting them, causing both to reflect on their times of loneliness. \\
         \bottomrule
    \end{tabular}
    \label{tab:scenarios}
\end{table}
\setlength{\tabcolsep}{6pt}

\subsection{Meta-emotions and Meta-relationships}
Tab. \ref{tab:meta-emotions} and Tab. \ref{tab:meta-relationships} show the correspondence between emotions and relationships as prompted to the actors, and our definition of meta-emotions and meta-relationships for more convenient visualization and analysis of our data.

\begin{table}[h]
    \caption{Meta-emotions and Corresponding Emotions}
    \centering
    \begin{tabular}{p{5cm}p{8cm}}
    \toprule
      Meta-emotion  & Emotion  \\
    \midrule
      \multirow{4}{*}{Surprise-related} & Surprise \\
      & Awe \\
      & Confusion \\
      & Excitement \\
    \midrule
      \multirow{4}{*}{Fear-related} & Anxiety \\
      & Fear \\
      & Horror \\
      & Disgust \\
    \midrule
      \multirow{6}{*}{Social} & Admiration \\
      & Adoration \\
      & Romance \\
      & Interest \\
      & Awkwardness \\
      & Boredom \\
    \midrule
      \multirow{7}{*}{Self-focused} & Satisfaction \\
      & Relief \\
      & Craving \\
      & Nostalgia \\
      & Entrancement \\
      & Calmness \\
      & Aesthetic appreciation \\
    \midrule
      \multirow{2}{*}{Primitive (Positive)} & Joy \\
      & Amusement \\
    \midrule
      \multirow{3}{*}{Primitive (Negative)} & Pain \\
      & Anger \\
      & Sadness \\
    \bottomrule
    \end{tabular}
    \label{tab:meta-emotions}
\end{table}

\begin{table}[h]
    \caption{Meta-relationships and Corresponding Relationships}
    \centering
    \begin{tabular}{p{5cm}p{8cm}}
    \toprule
      Meta-relationship  & Relationship  \\
    \midrule
      \multirow{6}{*}{Family} & Siblings \\
      & Parent / Child \\
      & Grandparent / Child \\
      & Cousins \\
      & In-Laws \\
      & Romantic Partner \\
    \midrule
      \multirow{6}{*}{Friends} & Friends With Similar Hobbies \\
      & Childhood Friends \\
      & Commuting Buddies \\
      & Sports Teammates \\
      & Interest Group Members \\
      & Churchmates \\
    \midrule
      \multirow{4}{*}{Work} & Co-workers \\
      & Business Partners \\
      & Boss / Subordinate \\
      & Mentor / Mentee \\
    \midrule
      \multirow{5}{*}{School} & Classmates \\
      & Senior Student / Junior Student \\
      & Bully / Classmate \\
      & Teacher / Student \\
      & Principal / Student \\
    \midrule
      \multirow{2}{*}{Professional} & Doctor / Patient \\
      & Waiter (Waitress) / Diner \\
    \bottomrule
    \end{tabular}
    \label{tab:meta-relationships}
\end{table}

\subsection{Capture Details}
Actors were found through referral or contacted on a local acting Facebook group. Previous acting experience and basic English proficiency were required. After an initial screening, we further narrowed our participants down by a multitude of criteria, including the quality and nature of previous acting, spoken proficiency, dataset diversity objectives, and availability on motion capture schedules. The final roster of participants all had prior acting experience such as drama, TV, or films, but not all had previous experience in impromptu acting. For each actor, we first recorded a mesh registration video using the Live Link Face app \cite{EpicGames_LiveLinkFacialCapture}, containing footage of the actor facing directly at the camera while moving their head side to side. This video was then processed in Unreal Engine to be exported as the full head mesh, which contained 24049 vertices. To perform the capture, we recorded the MoCap with 120 frames per second and the facial RGB-D video with 60 frames per second. We used Radical Variance FaceCam as the MoCap helmet, RODE HS2 as the primary microphone connected via two adapters in series to the iPhone, TRS-to-TRRS, then TRRS-to-Lightning. DJI Mic was used as the secondary microphone, for backup and real-time monitoring. The Vicon Lock Studio onboard timecode was physically synced to a Tentacle Sync E, then wirelessly broadcasted via a Bluetooth-based syncing protocol to the capture iPhones.

\subsection{Facial Mesh to ARKit Conversion}
To convert an input facial mesh sequence into ARKit blendshape parameters, we employ a least-squares fitting approach. First, the input mesh sequence is represented as a series of per-frame deformations by subtracting a neutral facial mesh. Similarly, the target ARKit blendshapes are formulated as delta shapes relative to a neutral expression. Using \verb|scipy.sparse.linalg.lsqr| , we then solve a linear least-squares problem for each frame in the input sequence to determine the optimal set of blendshape weights that, when applied to the ARKit delta blendshapes, best reconstruct the input frame's deformation. This process effectively translates the facial animation into the ARKit format.

\begin{table}[t]
\caption{Contact between actor's hands and own body parts}
\resizebox{\textwidth}{!}{%
\begin{tabular}{rl|ll|ll|ll|ll}
\hline
\multicolumn{2}{c|}{\multirow{2}{*}{\textbf{Self-contacts}}}                                                                                              & \multicolumn{2}{c|}{\textbf{Hand-head}}                            & \multicolumn{2}{c|}{\textbf{Hand-shoulder}}                        & \multicolumn{2}{c|}{\textbf{Hand-elbow}}                           & \multicolumn{2}{c}{\textbf{Hand-hand}}                             \\
\multicolumn{2}{c|}{}                                                                                                                                     & Dur. (s) & \begin{tabular}[c]{@{}l@{}}Rate\\ (min/hr)\end{tabular} & Dur. (s) & \begin{tabular}[c]{@{}l@{}}Rate\\ (min/hr)\end{tabular} & Dur. (s) & \begin{tabular}[c]{@{}l@{}}Rate\\ (min/hr)\end{tabular} & Dur. (s) & \begin{tabular}[c]{@{}l@{}}Rate\\ (min/hr)\end{tabular} \\ \hline
\multirow{2}{*}{\textbf{\begin{tabular}[c]{@{}r@{}}By\\ gender\end{tabular}}}              & Male                                                         & 27       & 0.1                                                     & 0        & 0.0                                                     & 45       & 0.2                                                     & 2206     & 7.3                                                     \\
                                                                                           & Female                                                       & 15       & 0.1                                                     & 63       & 0.2                                                     & 73       & 0.2                                                     & 2239     & 7.4                                                     \\ \hline
\multirow{5}{*}{\textbf{\begin{tabular}[c]{@{}r@{}}By meta-\\ relationships\end{tabular}}} & Family                                                       & 9        & 0.0                                                     & 57       & 0.2                                                     & 37       & 0.1                                                     & 1892     & 6.6                                                     \\
                                                                                           & Friends                                                      & 1        & 0.1                                                     & 2        & 0.1                                                     & 3        & 0.1                                                     & 624      & \textbf{22.8}                                           \\
                                                                                           & Professional                                                 & 5        & 0.0                                                     & 3        & 0.0                                                     & 54       & 0.4                                                     & 1107     & 9.0                                                     \\
                                                                                           & School                                                       & 0        & 0.0                                                     & 0        & 0.0                                                     & 20       & 1.1                                                     & 322      & \textbf{17.6}                                           \\
                                                                                           & Work                                                         & 25       & 0.3                                                     & 0        & 0.0                                                     & 2        & 0.0                                                     & 308      & 4.1                                                     \\ \hline
\multirow{6}{*}{\textbf{\begin{tabular}[c]{@{}r@{}}By meta-\\ emotions\end{tabular}}}      & Fear-related                                                 & 1        & 0.0                                                     & 22       & 0.2                                                     & 57       & \textbf{0.6}                                            & 489      & 5.0                                                     \\
                                                                                           & \begin{tabular}[c]{@{}l@{}}Primitive\\ negative\end{tabular} & 7        & 0.1                                                     & 3        & 0.0                                                     & 13       & 0.2                                                     & 590      & 7.7                                                     \\
                                                                                           & \begin{tabular}[c]{@{}l@{}}Primitive\\ positive\end{tabular} & 0        & 0.0                                                     & 0        & 0.0                                                     & 0        & 0.0                                                     & 245      & 5.1                                                     \\
                                                                                           & Self-focused                                                 & 6        & 0.0                                                     & 0        & 0.0                                                     & 10       & 0.1                                                     & 1058     & 7.2                                                     \\
                                                                                           & Social                                                       & 6        & 0.0                                                     & 1        & 0.0                                                     & 36       & 0.3                                                     & 1551     & \textbf{11.4}                                           \\
                                                                                           & \begin{tabular}[c]{@{}l@{}}Surprise-\\ related\end{tabular}  & 20       & 0.2                                                     & 36       & 0.4                                                     & 0        & 0.0                                                     & 510      & 5.2                                                     \\ \hline
\textbf{Aggregate}                                                                         & Total                                                        & 42       & 0.1                                                     & 64       & 0.1                                                     & 119      & 0.2                                                     & 4446     & 7.3                                                     \\ \hline
\end{tabular}%
}
\hspace{1em}
\label{tab:self-contacts}
\end{table}

\begin{table}[t]
\caption{Contact between actor's hands and partner's body parts}
\resizebox{\textwidth}{!}{%
\begin{tabular}{rl|ll|ll|ll|ll}
\hline
\multicolumn{2}{c|}{\multirow{2}{*}{\textbf{Cross-contacts}}}                                                                                             & \multicolumn{2}{c|}{\textbf{Hand-head}}                            & \multicolumn{2}{c|}{\textbf{Hand-shoulder}}                        & \multicolumn{2}{c|}{\textbf{Hand-elbow}}                           & \multicolumn{2}{c}{\textbf{Hand-hand}}                             \\
\multicolumn{2}{c|}{}                                                                                                                                     & Dur. (s) & \begin{tabular}[c]{@{}l@{}}Rate\\ (min/hr)\end{tabular} & Dur. (s) & \begin{tabular}[c]{@{}l@{}}Rate\\ (min/hr)\end{tabular} & Dur. (s) & \begin{tabular}[c]{@{}l@{}}Rate\\ (min/hr)\end{tabular} & Dur. (s) & \begin{tabular}[c]{@{}l@{}}Rate\\ (min/hr)\end{tabular} \\ \hline
\multirow{2}{*}{\textbf{\begin{tabular}[c]{@{}r@{}}By\\ gender\end{tabular}}}              & Male                                                         & 0        & 0.0                                                     & 32       & 0.1                                                     & 77       & 0.3                                                     & 193      & 0.6                                                     \\
                                                                                           & Female                                                       & 0        & 0.0                                                     & 3        & 0.0                                                     & 26       & 0.1                                                     & 193      & 0.6                                                     \\ \hline
\multirow{5}{*}{\textbf{\begin{tabular}[c]{@{}r@{}}By meta-\\ relationships\end{tabular}}} & Family                                                       & 0      & 0.0                                                     & 19       & 0.1                                                     & 102      & \textbf{0.4}                                            & 180      & \textbf{0.6}                                            \\
                                                                                           & Friends                                                      & 0      & 0.0                                                     & 0      & 0.0                                                     & 0      & 0.0                                                     & 0        & 0.0                                                     \\
                                                                                           & Professional                                                 & 0        & 0.0                                                     & 9        & 0.1                                                     & 0        & 0.0                                                     & 2        & 0.0                                                     \\
                                                                                           & School                                                       & 0        & 0.0                                                     & 0        & 0.0                                                     & 0        & 0.0                                                     & 0        & 0.0                                                     \\
                                                                                           & Work                                                         & 0        & 0.0                                                     & 0        & 0.0                                                     & 1        & 0.0                                                     & 8        & 0.1                                                     \\ \hline
\multirow{6}{*}{\textbf{\begin{tabular}[c]{@{}r@{}}By meta-\\ emotions\end{tabular}}}      & Fear-related                                                 & 0        & 0.0                                                     & 2        & 0.0                                                     & 50       & \textbf{0.5}                                            & 47       & 0.5                                                     \\
                                                                                           & \begin{tabular}[c]{@{}l@{}}Primitive\\ negative\end{tabular} & 0        & 0.0                                                     & 17       & 0.2                                                     & 35       & \textbf{0.5}                                            & 35       & 0.5                                                     \\
                                                                                           & \begin{tabular}[c]{@{}l@{}}Primitive\\ positive\end{tabular} & 0        & 0.0                                                     & 3        & 0.1                                                     & 1        & 0.0                                                     & 27       & 0.6                                                     \\
                                                                                           & Self-focused                                                 & 0        & 0.0                                                     & 9        & 0.1                                                     & 15       & 0.1                                                     & 62       & 0.4                                                     \\
                                                                                           & Social                                                       & 0        & 0.0                                                     & 0        & 0.0                                                     & 0        & 0.0                                                     & 10       & 0.1                                                     \\
                                                                                           & \begin{tabular}[c]{@{}l@{}}Surprise-\\ related\end{tabular}  & 0        & 0.0                                                     & 2        & 0.0                                                     & 1        & 0.0                                                     & 9        & 0.1                                                     \\ \hline
\textbf{Aggregate}                                                                         & Total                                                        & 0        & 0.0                                                     & 35       & 0.1                                                     & 104      & 0.2                                                     & 193      & 0.3                                                     \\ \hline
\end{tabular}%
}
\hspace{1em}
\label{tab:cross-contacts}
\end{table}

\subsection{Details on Self-contacts and Cross-contacts}
For each contact type (hand-head, hand-shoulder, hand-elbow, hand-hand), the total duration of contact in seconds, and the rate of occurrence in minutes per hour of a given scenario type is computed. For hand-head, contact is defined as left and/or right hand touching one's head. For hand-shoulder, contact is defined as either left or right hand touching either left or right shoulder. For hand-elbow, self-contact is defined as either left hand touching right elbow, or right hand touching left elbow, and cross-contact by either left or right hand touching their partner's left or right elbow. Finally, for hand-hand, self-contact is defined as both hands touching, and cross-contact by either left or right hand touching their partner's left or right hand.

\section{Network Architecture}

\subsection{Training Details}
We split all 241 sequences of InterAct into a training set and a testing set. The training set contains 208 sequences (86.31\%), while the testing set contains the remaining 33 sequences (13.69\%). Our body models are trained on 5-second segments (150 frames)  evenly cut from the whole initial sequences and tested on 10 second segments (300 frames). For the body motion generation, both networks for major joints and remaining joints in our method are trained for 60 epochs, each epoch taking around 1.5 hours on an NVIDIA A100 GPU. The batch size is set to 32, and the Adam optimizer with a learning rate of 3e-4 is used to train the network. We empirically find that with longer training time the generated motions become smoother, while the accuracy of the motions decrease. We obtain the code for EMAGE \cite{liu2023emage} and the one in \cite{ng2024audio2photoreal} from their Github repositories, then train and test their models on InterAct with the same splits. 

For the face model, blendshape parameter sequences are cut into 5-second segments (150 frames) for training. At test time, the full, unaltered sequences from the dataset are used for inference. The Periodic Positional Encoding \cite{fan2022faceformer} is set to a period of 30 frames, i.e. 1 second. For both the base and fine-tuned model, we use a batch size of 2 and a learning rate of 1e-6, which yields a predictably smooth loss curve. For the base model, the network is trained for 3429 epochs, with each epoch taking around 3 minutes on an NVIDIA A100 GPU. For the fine-tuned model, the final network is additionally trained for 5 more epochs on our small lip-accurate dataset. We find that fine-tuning for 5 to 40 epochs yields acceptable results, and fine-tuning for longer tends to trade off emotional expressiveness for better lip accuracy.

\subsection{Face Model Fine-Tuning}
While domain specific performance may improve during fine-tuning, this process may also introduce adverse changes to other parts of the network, weakening performance outside the fine-tuning dataset. In the case of our face model, we notice that with simple fine-tuning, the lip shape of all identities would eventually morph to that of the actor in our fine-tuning dataset. A method we discovered to be effective in our architecture is to only swap the denoiser weights of the base model with the fine-tuned model while keeping other layers untouched. We illustrate the effect of this lip-shape preserving property in Fig. \ref{fig:face_comp}, and in the supplementary video.

\begin{figure}
    \centering
    \includegraphics[width=1.0\linewidth]{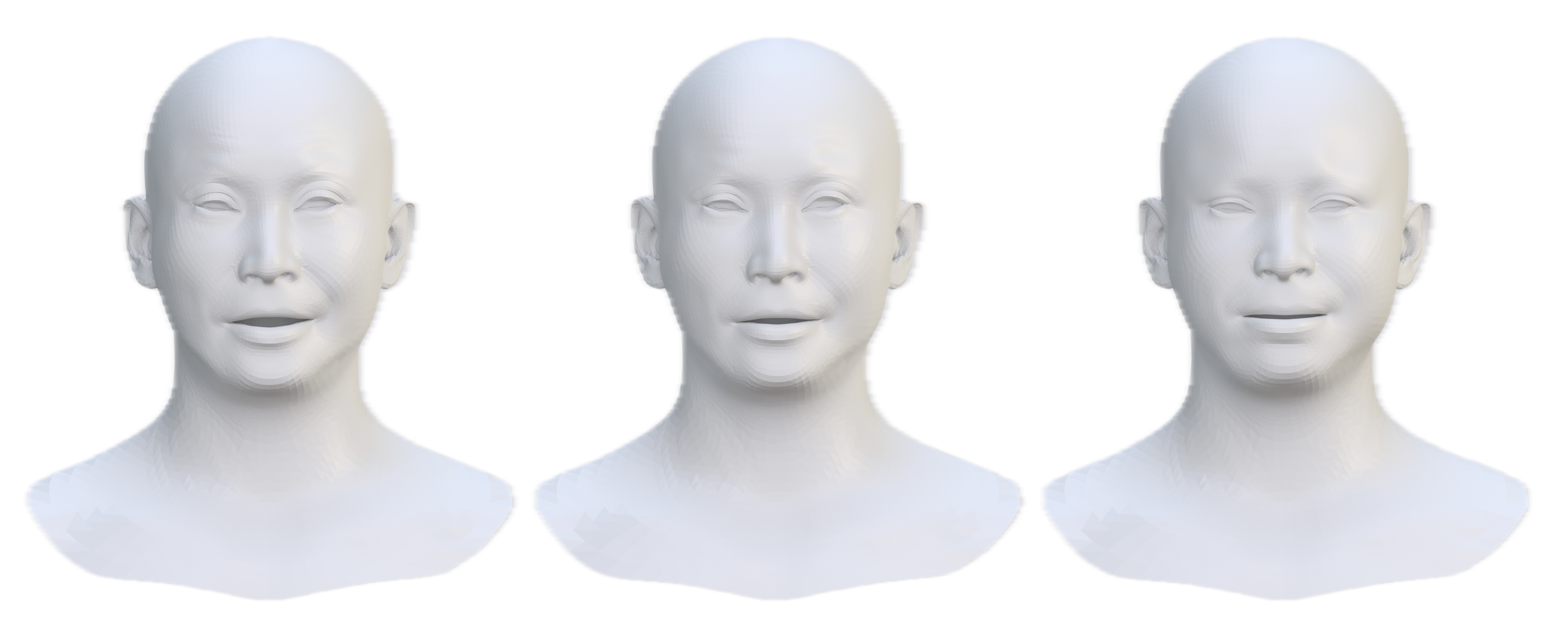}
    \caption{Left: Base model, Middle: Final model with denoiser weight injection, Right: Fine-tuned model. All three heads are showing the same frame given an input audio. Note how the base model tends to open the mouth wider, sacrificing lip accuracy, while the fine-tuned model sacrifices the actor's unique lip shape. The middle model achieves a balance between lip accuracy, maintaining lip shape, and emotional expressiveness.}
    \label{fig:face_comp}
\end{figure}

\section{User Study}
\begin{figure}
    \centering
    \includegraphics[width=0.7\linewidth, trim={0.3cm 0cm 0.3cm 0.7cm}, clip]{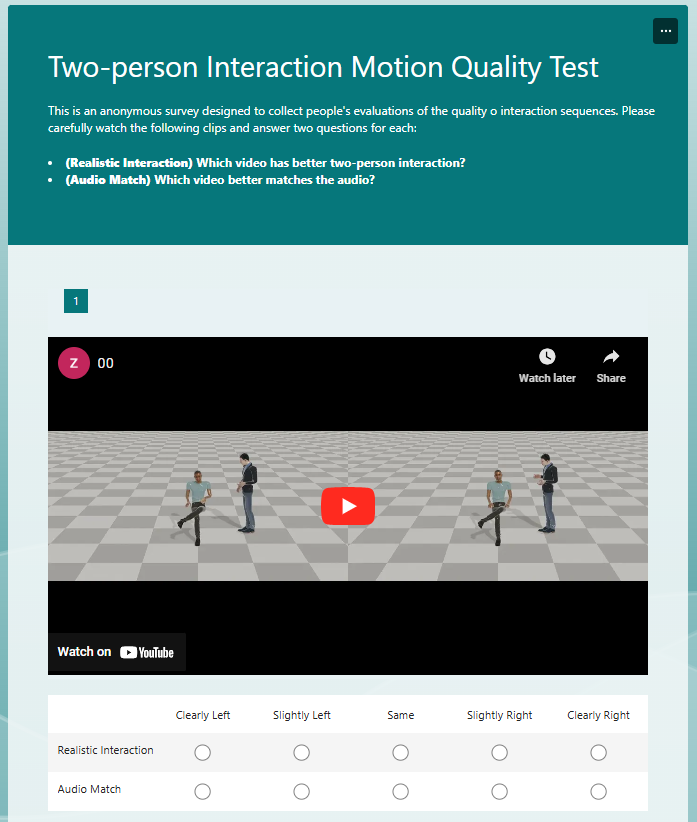}  
    \caption{Interface of qualitative study for body motions. }    
    \label{fig:user_study_body}
\end{figure}

We conduct user studies on results generated by our method and other baselines for subjective evaluations.

\subsection{Face} For the face, we randomly select 5 sequences ranging from 47 seconds to 1 minute 30 seconds from our test set to render a side-by-side comparison between our methods and FaceFormer. For each sequence, each of our participants then provides a realism score for every method, ranging from 1 (Worst) to 5 (Best). In total, 20 people participated in the user study for facial motions.

\subsection{Body} For the body part, following LDA \cite{alexanderson2023listen}, we pick one 10-second segment from each of 33 testing sequences and render the body motions synthesized by different methods. The baseline method to compare against ours is $\text{LDA}_2$, where all the joints of both actors are estimated via one single network. The 33 segments are carefully selected to include various relationships, emotions and scenarios to better represent the whole testing set. Then, an anonymous evaluator is shown several pairs of rendered videos from our method and the baseline. The evaluator is asked to rate the shown videos based on how realistic the interaction is and also how well the body motions match the speech. The interface is shown in Figure \ref{fig:user_study_body}. Altogether, 38 people take part in the user study for body motions.

\end{document}